# Effects of Deep Learning Architecture and Pretraining on Automated Vocal Fold Segmentation in Ex Vivo High-Speed Phonation Images

Varun Vinayak Kalaiarasan [a], Charles Farbos de Luzan [a, b], Khalid Malik [b], Ephraim Gutmark [c], Liran Oren [a, b]

[a] Department of Biomedical Engineering, University of Cincinnati, Cincinnati, OH, 45219

[b] Department of Otolaryngology, University of Cincinnati, Cincinnati, OH, 45267

[c] Department of Aerospace Engineering & Engineering Mechanics, University of Cincinnati, Cincinnati, OH, 45221

## Abstract

Two-dimensional (2D) intraglottal particle image velocimetry (PIV) of *ex vivo* phonation experiments quantifies vocal fold medial surface dynamics and the intraglottal flow field. However, it relies on manual segmentation because residual glare, low tissue-to-flow contrast, and tracer-particle speckle obscure the vocal fold boundaries. This study assessed how deep learning architecture and ImageNet pretraining affect automated vocal fold segmentation in these high-speed images. Four established architectures from three categories were benchmarked: the convolutional neural network (CNN)-based UNet and UNet++, the hybrid CNN-transformer-based TransUNet, and the pure transformer-based Swin-UNet, the latter two trained from scratch and from ImageNet-pretrained weights. Using 16,344 expert-annotated images from 227 phonation trials of excised canine larynges, the six pipelines were trained with five seeds under a standardized protocol and compared on 50 held-out test trials using trial-level Wilcoxon signed-rank tests with Holm correction. Mean Dice similarity coefficient (DSC) spanned only 0.9734–0.9764, so the pipelines differed mainly in boundary fidelity. Pretraining significantly improved the boundary F1 score (BF1) and 95$^{th}$ percentile Hausdorff distance (HD95) of the TransUNet and Swin-UNet, about four times more for the Swin-UNet. Once pretrained, the two performed similarly, although the Swin-UNet has only about a quarter of the trainable parameters. Without pretraining, the TransUNet had a significantly higher BF1 and a lower HD95 than both CNN-based pipelines and about half their share of images with HD95 $>$ 50 pixels. To our knowledge, this is the first study to apply deep learning to vocal fold segmentation in 2D intraglottal PIV, and it offers a reference framework for segmenting experimental flow-visualization images with noisy optical conditions or highly deformable objects.

## 1. Introduction

Voice production is a complex phenomenon governed by three interlinked components: subglottal pressure from the lungs, sound generation at the vocal folds, and resonance within the vocal tract [1]. According to the source-filter theory, the primary sound source is generated at the glottis, where vibration of the vocal folds modulates the airflow from the lungs, a process known as phonation [2]. During phonation, airflow from the lungs interacts with the vocal folds, exciting them into self-sustained oscillation [1]. This fluid-structure interaction between the medial surface of the vocal folds and the intraglottal flow governs how efficiently aerodynamic energy from the lungs is converted into acoustic energy, a ratio termed vocal efficiency (VE) [3].

During vibration, the medial surface of the vocal folds moves continuously, with the glottis cyclically alternating between convergent and divergent shapes [4]. Importantly, the divergent shape of the glottis during the closing phase of vibration causes the intraglottal flow to separate from the medial surface of the vocal folds, producing negative pressures that accelerate glottal closure [5–7]. This accelerated closure is associated with increased acoustic intensity and VE [8,9]. These fundamental and mechanistic insights were observed in computational simulations [6,7], and validated through *ex vivo* phonation experiments using excised canine larynges in conjunction with two-dimensional (2D) intraglottal particle image velocimetry (PIV) in the mid-coronal plane of the vocal folds [5,9].

Briefly, 2D PIV is a flow visualization technique in which a high-repetition-rate planar laser sheet illuminates tracer particles seeded in the flow while a high-speed camera acquires images synchronized with the laser pulses [10]. Intraglottal PIV thereby captures medial surface dynamics and the intraglottal flow field simultaneously during vocal fold vibration [11]. These *ex vivo* phonation experiments have advanced the understanding of healthy and pathological voice and have generated treatment-relevant insights for severe voice disorders such as unilateral vocal fold paralysis [12,13].

To quantitatively analyze vocal fold medial surface dynamics and the intraglottal flow field, the vocal folds must first be delineated by a binary mask. High-speed imaging generates a large volume of data, on the order of tens of thousands of frames, and the optical conditions required for intraglottal PIV produce residual laser glare, low tissue-to-flow contrast, and tracer-particle speckle that obscure the vocal fold boundaries. Given these constraints, binary masks are currently generated through manual segmentation. This process is labor- and time-intensive, prone to user bias, and unsuitable for large-scale comparative studies. Manual segmentation therefore is a substantial bottleneck, inhibiting the translation from data acquisition to physiological insight. Without automated segmentation, the number of laryngeal specimens, simulated interventions, and phonation conditions that can be investigated is bounded by annotation effort.

Several approaches have been developed to automate the segmentation of PIV images, but each relies on an assumption that intraglottal PIV does not satisfy. Masking based on pixel-intensity statistics identifies static image regions from the temporal variation in intensity across an image sequence [14], whereas the vocal folds move continuously throughout the vibratory cycle, so pixels near the medial surface alternate between tissue and seeded flow. Photogrammetric masking locates rigid moving bodies [15] and may struggle with tissue that deforms throughout each cycle. Fluorescence-based ray-casting tracks moving and deformable objects but it requires the object to be coated with fluorescent paint before data acquisition [16], which conflicts with the experimental protocol needed to minimize residual laser glare. Deep learning models are not bound by these assumptions, and a convolutional autoencoder was used to mask semi-transparent, moving, and deforming objects in PIV images, including a swimming jellyfish [17]. This proof of concept showed that learned masks are feasible for PIV, but it relied on a relatively small dataset and did not report comprehensive metrics quantifying segmentation performance.

In laryngology specifically, a considerable body of work, supported by publicly available annotated datasets [18], has focused on utilizing deep learning to segment the glottis and the vocal folds. Examples include ensemble models [19], the Segment Anything Model prompted by UNet-generated glottis masks [20], and a standalone UNet-based architecture [21], showing that established biomedical semantic segmentation architectures can be adapted to laryngeal segmentation tasks. These models, however, were evaluated on clinical laryngeal endoscopic images, which view the vocal folds from above and are not subject to the same noisy optical environment of intraglottal PIV. Consequently, how architecture affects segmentation performance under these experimental conditions has yet to be investigated.

Beyond laryngology, deep learning is widely used to segment tumors, organs, and other tissues [22], and the semantic segmentation architectures developed for these tasks differ in how they capture spatial context. The UNet is considered a convolutional neural network (CNN) baseline, designed to perform well with limited training data [23], and the UNet++ redesigns its skip connections to narrow the semantic gap between encoder and decoder features [24]. Convolution gives CNNs a strong inductive bias toward local features, but the effective receptive field of a CNN covers only a fraction of its theoretical receptive field [25]. As a result, distant image regions contribute little to each output pixel. Hybrid CNN-transformer- and pure transformer-based architectures instead use self-attention to model long-range relationships within an image [26], which could mitigate this limitation [27]. The TransUNet combines a CNN encoder with a transformer encoder [28], whereas the Swin-UNet is built entirely from transformer blocks that compute self-attention within shifted local windows [29]. Transformers, however, lack the locality and translation equivariance built into convolution and tend to generalize poorly when training data are limited [26].

Benchmarking studies have been used to assess how established semantic segmentation architectures perform on new datasets and imaging domains [30,31]. For hybrid CNN-transformer and pure transformer architectures, benchmarking must also account for pretraining, because transformers necessitate large-scale pretraining on natural images to compensate for their data dependency [26]. In biomedical image classification,

ImageNet pretraining closed the performance gap between vision transformers (ViTs) and CNNs that was present when the ViTs were trained from scratch [32]. Furthermore, in a recent histopathological benchmark, hybrid CNN-transformer and pure transformer architectures gained more from pretraining than CNNs [33]. Intraglottal PIV high-speed images, however, are grayscale, laser-illuminated, and are speckled with tracer particles, so neither finding can be assumed to hold.

To our knowledge, this study is the first to apply deep learning to vocal fold segmentation in 2D intraglottal PIV high-speed images, and it does so by systematically benchmarking CNN-based (UNet and UNet++) [23,24], hybrid CNN-transformer-based (TransUNet) [28], and pure transformer-based (Swin-UNet) [29] semantic segmentation architectures with an emphasis on boundary fidelity. Accurate delineation of the vocal folds is critical, as analysis of both medial surface dynamics and the intraglottal flow field depends on where these boundaries lie. The main contributions of this study are: (1) a controlled comparison of four established architectures spanning three categories, trained under a standardized protocol on a novel dataset of 16,344 expert-annotated images from 227 *ex vivo* phonation trials; (2) a statistical assessment of the effect of architecture on vocal fold segmentation, with each model trained across five random seeds and contrasts tested at the trial level; (3) an equivalent assessment of whether ImageNet pretraining improves the hybrid CNN-transformer- and pure transformer-based architectures on images far removed from the natural images used for pretraining; and (4) a comprehensive evaluation of segmentation performance using regional overlap and boundary fidelity metrics, together with a characterization of where large errors occurred across the vibratory cycle.

Together, these contributions provide empirical guidance for automating vocal fold segmentation in intraglottal PIV. More broadly, they test established architectures and ImageNet pretraining on experimental flow-visualization images of rapidly deforming tissue with noisy optical conditions, far outside the domains in which these architectures were developed.

## 2. Methods

### 2.1 Overview

The main objective of this work is to systematically benchmark established CNN-, hybrid CNN-transformer-, and pure transformer-based semantic segmentation architectures, trained and evaluated under a standardized protocol, to delineate the vocal folds from 2D intraglottal PIV high-speed images. This automated approach aims to address the substantial limitations of the current manual segmentation process. An overview of the methodology is provided in Fig. 1.

### 2.2 Dataset and image acquisition

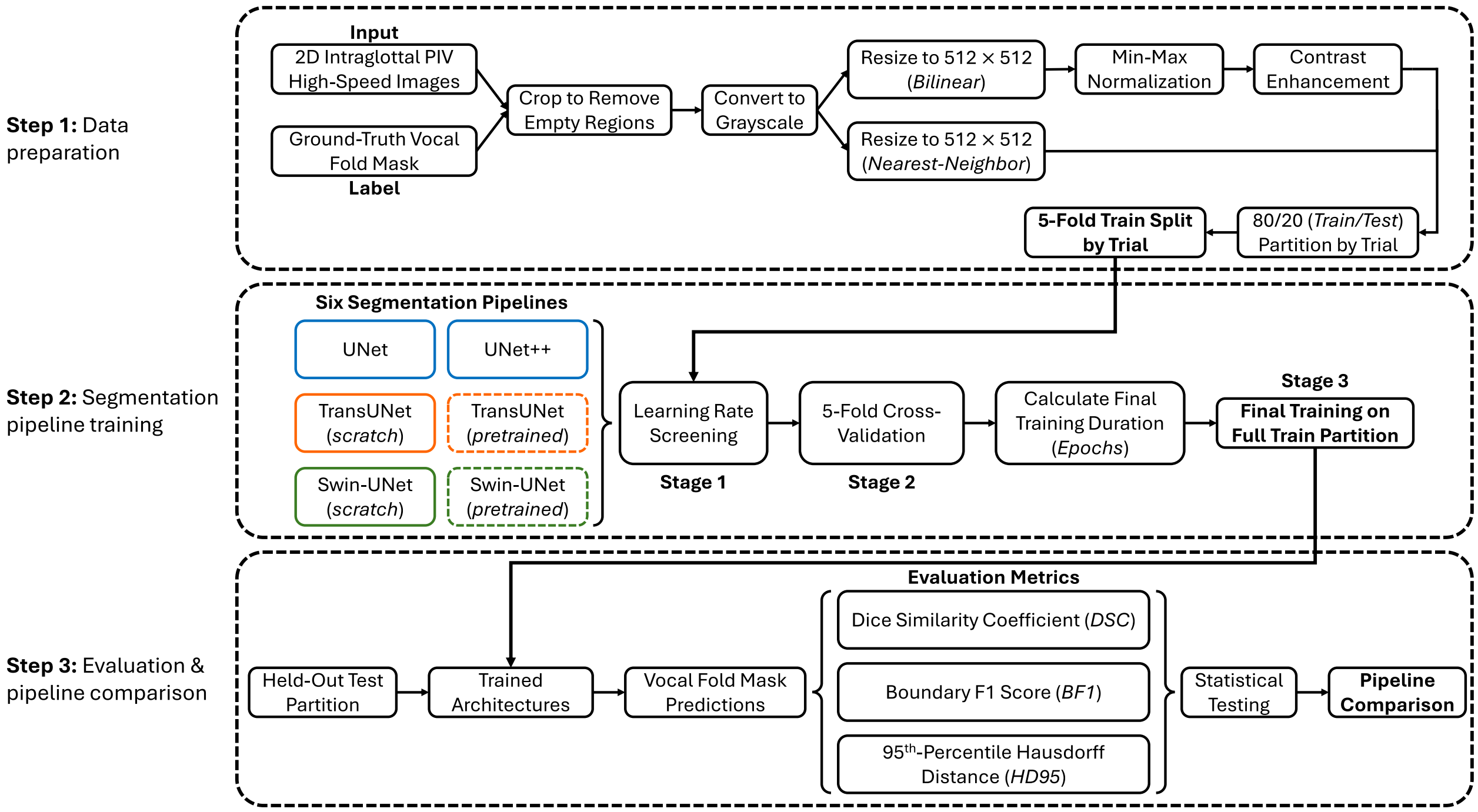

**Fig. 1.** Methodology for evaluating semantic segmentation architectures for vocal fold delineation from two-dimensional (2D) intraglottal particle image velocimetry (PIV) high-speed images. Step (1) prepares the dataset and partitions it by trial. Step (2) systematically trains six semantic segmentation pipelines using a standardized three-stage protocol. The architectures span three categories: convolutional neural network-based (CNN; blue), hybrid CNN-transformer-based (orange), and pure transformer-based (green). Step (3) quantitatively evaluates segmentation performance and performs trial-level statistical testing.

The dataset was obtained from previously conducted *ex vivo* phonation experiments using excised canine larynges. All animal procedures were approved by the University of Cincinnati Institutional Animal Care and Use Committee (IACUC protocol 26-04-14-02). The experimental preparation and 2D intraglottal PIV methodology have been reported previously [34,35]. Briefly, the excised larynges were subjected to self-sustained phonation, and a high-speed camera synchronized with a high-repetition-rate planar laser sheet was used to visualize the vocal fold medial surface dynamics and seeded intraglottal flow in the mid-coronal plane. The high-speed imaging was phase-locked to an electroglottography reference signal [35], with each vibratory cycle represented by 72 phase-resolved images acquired at 5° increments from 0° to 355°.

The optical conditions inherent to intraglottal PIV produced residual laser glare, low tissue-to-flow contrast, and tracer-particle speckle that obscured the vocal fold boundaries. Therefore, this noisy experimental environment required manual segmentation via a binary mask. These masks served as the ground truth for deep learning model training and evaluation. Representative imaging conditions and a corresponding ground-truth mask are shown in Fig. 2. In total, the dataset consisted of 8 experiments and 227 trials, comprising 16,344 images.

## 2.3 Ground-truth annotation protocol

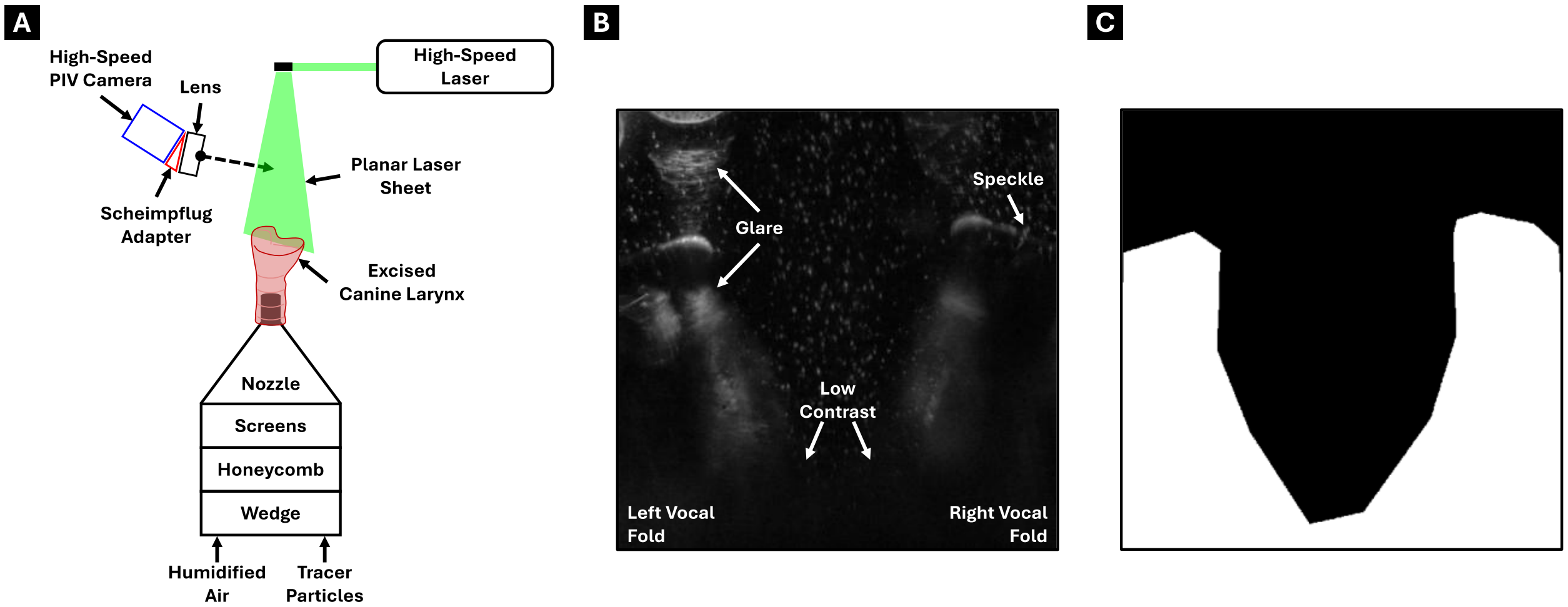

**Fig. 2.** *Ex vivo* two-dimensional (2D) intraglottal particle image velocimetry (PIV) experimental setup and representative image-mask pair. (A) Schematic of the experimental setup. (B) Representative high-speed image highlighting residual laser glare, low tissue-to-flow contrast, and tracer-particle speckle that obscure the vocal fold boundaries. (C) Corresponding manually generated binary ground-truth mask.

Delineation of the vocal folds served two purposes: to characterize medial surface dynamics and to quantify the intraglottal flow field by separating the seeded flow from the rapidly deforming laryngeal tissue. Manual segmentation was originally performed by subject-matter experts (SMEs) for PIV analysis, leveraging a custom-built application developed in MATLAB (MathWorks, Natick, MA). The annotation focused on the divergent phases, the remaining phases were annotated intermittently, and linear interpolation was used between the annotated phases. For this study, images with considerably erroneous masks were re-annotated by an SME using the same application, in some cases for an entire trial. Using a polygon-generation tool, eight sequential vertices were placed along the observed boundary of each vocal fold, yielding 16 vertices per image. The spacing between vertices was determined based on vocal fold morphology. The vertices were then connected and manually adjusted, with the enclosed region defining the binary mask. When an entire trial was re-annotated, manual segmentation began at 0°, and for subsequent phases, preceding and succeeding phase-resolved images were considered to maintain spatial and temporal continuity throughout the vibratory cycle.

## 2.4 Dataset preprocessing

The 2D intraglottal PIV high-speed images were acquired using DaVis software (LaVision GmbH, Göttingen, Germany). Because the images were collected across multiple experiments, they varied in spatial dimensions and often contained non-informative, empty regions. To standardize the data for deep learning, a preprocessing pipeline was developed in MATLAB (MathWorks, Natick, MA). Each image was first cropped to remove the empty regions and converted to grayscale. The PIV images and their corresponding ground-truth masks were then resized to 512 × 512 pixels, using bilinear interpolation for the PIV images and nearest-neighbor interpolation for the masks to preserve discrete boundary labels. This resolution provides a balance between computational efficiency and the preservation of spatial detail needed for semantic segmentation. PIV image intensities were normalized to the range [0, 1], and contrast-limited adaptive histogram equalization was applied only to the PIV images to enhance local contrast and improve the visibility of features relevant to segmentation. Prior to deep learning model input, the Python data pipeline enforced the expected grayscale format, 512 × 512 spatial dimensions, and normalized pixel intensities to ensure consistent input handling across all segmentation pipelines.

### 2.5 Dataset partitioning

As described in Section 2.2, the dataset comprised 16,344 phase-resolved high-speed images from 227 *ex vivo* phonation trials across 8 experiments. Each trial contained 72 images representing one vibratory cycle. Because phase-resolved images from the same trial are highly correlated, partitioning at the image level could distribute closely related observations across the training and held-out test sets and yield optimistically biased performance estimates [36]. To prevent this leakage, a nominal 80%/20% training/held-out test partition was generated at the trial level by randomizing trials independently within each experiment. All images from a given trial were therefore assigned exclusively to one partition. The resulting frozen split comprised 177 training trials (12,744 images) and 50 held-out test trials (3,600 images). The small deviation from an exact 80%/20% ratio resulted from independently partitioning whole trials within each experiment. The split was generated once, persisted, and reused across all six semantic segmentation pipelines. The held-out test set was reserved exclusively for final performance evaluation.

### 2.6 Semantic segmentation architectures

The PIV images and their corresponding manually generated ground-truth masks were used to train six semantic segmentation pipelines derived from four established architectures spanning three categories: CNN-based (UNet [23] and UNet++ [24]), hybrid CNN-transformer-based (TransUNet [28]), and pure transformer-based (Swin-UNet [29]). The TransUNet and Swin-UNet were evaluated from scratch and with ImageNet-pretrained initialization, yielding six pipelines: UNet, UNet++, TransUNet (*scratch*), TransUNet (*pretrained*), Swin-UNet (*scratch*), and Swin-UNet (*pretrained*). This design enabled comparison across architectural categories and, for the transformer-based architectures, assessment of the effect of pretraining.

#### 2.6.1 UNet

The CNN-based architectures evaluated in this study were the UNet and UNet++, whose schematics are presented in Fig. 3. The UNet is a widely established baseline for biomedical image segmentation and consists of a symmetric U-shaped encoder-decoder structure linked by skip connections [23], as shown in Fig. 3A. The encoder progressively reduces spatial resolution while increasing feature-channel depth, whereas the decoder restores spatial resolution and combines its features with the corresponding high-resolution encoder features through concatenation.

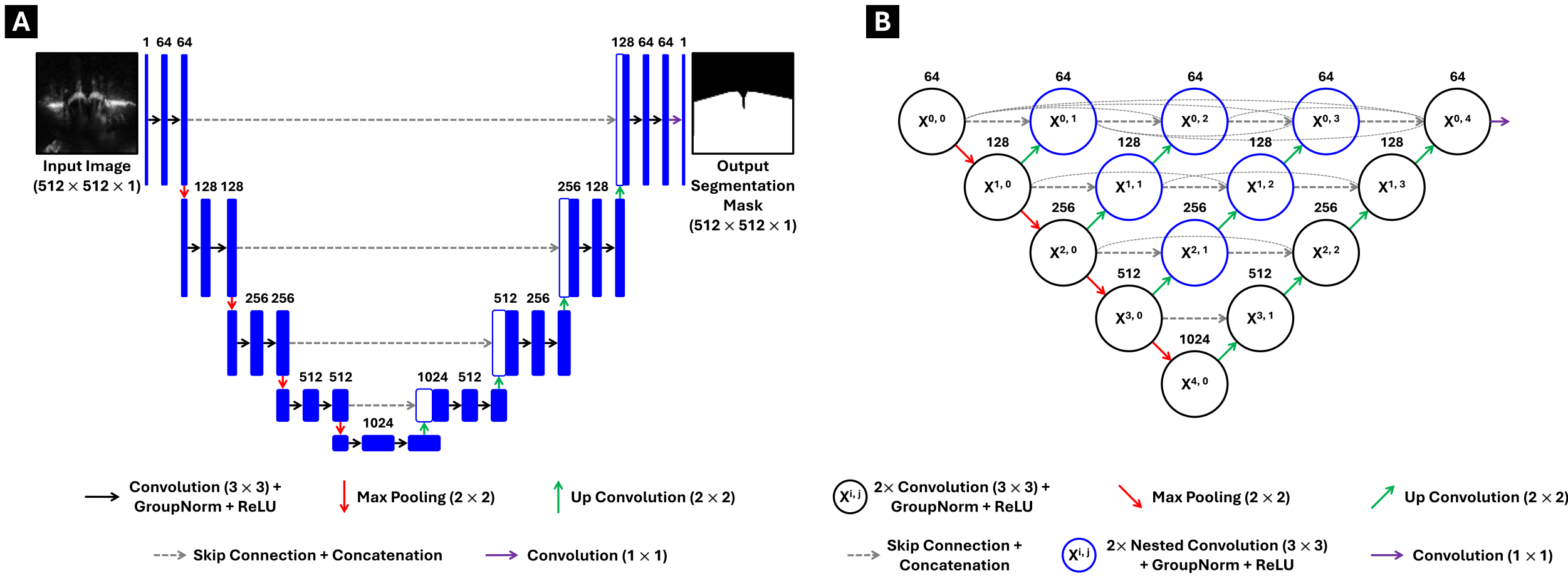

**Fig. 3.** Convolutional neural network (CNN)-based semantic segmentation architectures implemented to predict binary masks of the vocal folds from two-dimensional (2D) intraglottal particle image velocimetry (PIV) high-speed images. (A) UNet and (B) UNet++.

The UNet implementation accepted a single-channel 512 × 512 input and contained four encoder levels with channel widths of 64, 128, 256, and 512, followed by a 1024-channel bottleneck. The main building block, $\mathrm{H}(\cdot)$, consisted of two successive units, each containing a 3 × 3 convolution (padding = 1), followed by group normalization (GN; groups = 8) and a rectified linear unit (ReLU) activation. Each encoder level applied $\mathrm{H}(\cdot)$, followed by a 2 × 2 max-pooling operation that halved the spatial resolution. In the decoder, each level utilized a 2 × 2 up-convolution (transposed convolution) $\mathrm{U}(\cdot)$ to double the spatial resolution and halve the channel dimension. This was followed by concatenation with features from the corresponding encoder level and subsequent processing by $\mathrm{H}(\cdot)$. A final 1 × 1 convolution transformed the 64-channel decoder output into a single-channel 512 × 512 raw logit map. Convolutional and transposed convolutional weights were initialized using the Kaiming normal method, whereas the final 1 × 1 segmentation head was initialized using Xavier normal initialization. Biases, where present, were initialized to zero.

Several modifications were made to the original UNet implementation. All 3 × 3 convolutions used padding = 1 to preserve spatial dimensions and eliminate the need to crop encoder features before concatenation. GN was added instead of batch normalization (BN) to provide normalization independent of batch statistics. Spatial dropout (Dropout2d; rate = 0.15) was applied at the fourth encoder level after the final ReLU activation but before max pooling.

### 2.6.2 UNet++

A limitation of the UNet addressed by UNet++ is the semantic dissimilarity between the high-resolution encoder features and the more abstract decoder features due to direct concatenation through skip connections. To reduce this semantic gap, UNet++ maintains the UNet encoder-decoder backbone while replacing each direct skip connection with a series of nested, densely connected convolutional blocks [24], as illustrated in Fig. 3B. Given $\mathrm{H}(\cdot)$ and $\mathrm{U}(\cdot)$, let $x^{i,j}$ be the output of node $\mathrm{X}^{i,j}$, where $i$ indexes the encoder downsampling level and $j$ indexes the position of the node along its skip pathway. The output of each node is given by (1).

$$x^{i,j} = \begin{cases} \mathrm{H}\left(x^{i-1,j}\right), & j = 0 \\ \mathrm{H}\left(\left[x^{(i,0)}, \dots, x^{(i,j-1)}, \mathrm{U}\left(x^{(i+1,j-1)}\right)\right]\right), & j > 0 \end{cases} \tag{1}$$

Where $[\cdot]$ denotes concatenation along the channel dimension. Nodes at $j = 0$ constitute the original UNet encoder pathway and receive a single max-pooled input from the preceding encoder level. Nodes at $j > 0$ receive the outputs of all preceding nodes along the same horizontal skip pathway together with the upsampled output of the node located one level deeper.

The implemented UNet++ retained the UNet encoder channel widths of 64, 128, 256, and 512, followed by a 1024-channel bottleneck, and formed the full 15-node nested grid with 10 learned 2 × 2 transposed-convolution upsamplers. Each node used an independently parameterized instance of the same $\mathrm{H}(\cdot)$ building block, and the UNet spatial dropout configuration was preserved, with Dropout2d (rate = 0.15) applied at node $\mathrm{X}^{3,0}$. Deep supervision was disabled for consistency with the single-output design of the other segmentation pipelines. Instead, a 1 × 1 convolution was applied to the final node $\mathrm{X}^{0,4}$ to produce a single-channel raw logit map. Weight initialization and bias handling matched the UNet.

### 2.6.3 TransUNet

While CNN-based architectures provide strong spatial inductive biases, their local operations can limit the explicit modeling of long-range relationships within an image [25]. Self-attention provides a complementary mechanism for modeling non-local interactions [26], which may be beneficial for delineating the vocal folds from the intraglottal flow in PIV high-speed images. Therefore, this study evaluated the hybrid CNN-transformer-based TransUNet and the pure transformer-based Swin-UNet, whose schematics are presented in Fig. 4.

The TransUNet combines a CNN encoder with a transformer encoder [28], as displayed in Fig. 4A. The CNN encoder was a big transfer (BiT)-style ResNetV2 backbone with GN (groups = 32) and weight-standardized convolutions [37,38], in which the fourth stage was omitted and the remaining three stages contained 3, 4, and 9 residual blocks. Encoder features at 1/2, 1/4, and 1/8 of the input resolution, containing 64, 256, and 512 channels, respectively, were retained as skip connections. The resulting 32 × 32, 1024-channel feature map was projected by a 1 × 1 convolution to 768-dimensional embeddings, yielding 1024 tokens to which learned position embeddings were added to form the initial sequence $z_0$. The transformer encoder followed the ViT-B/16 configuration and contained $\mathrm{L} = 12$ layers [26], each comprising multi-head self-attention (MSA) and a multilayer perceptron (MLP) with Gaussian error linear unit (GELU) activation. The output of the $l$-th layer was defined by (2) and (3).

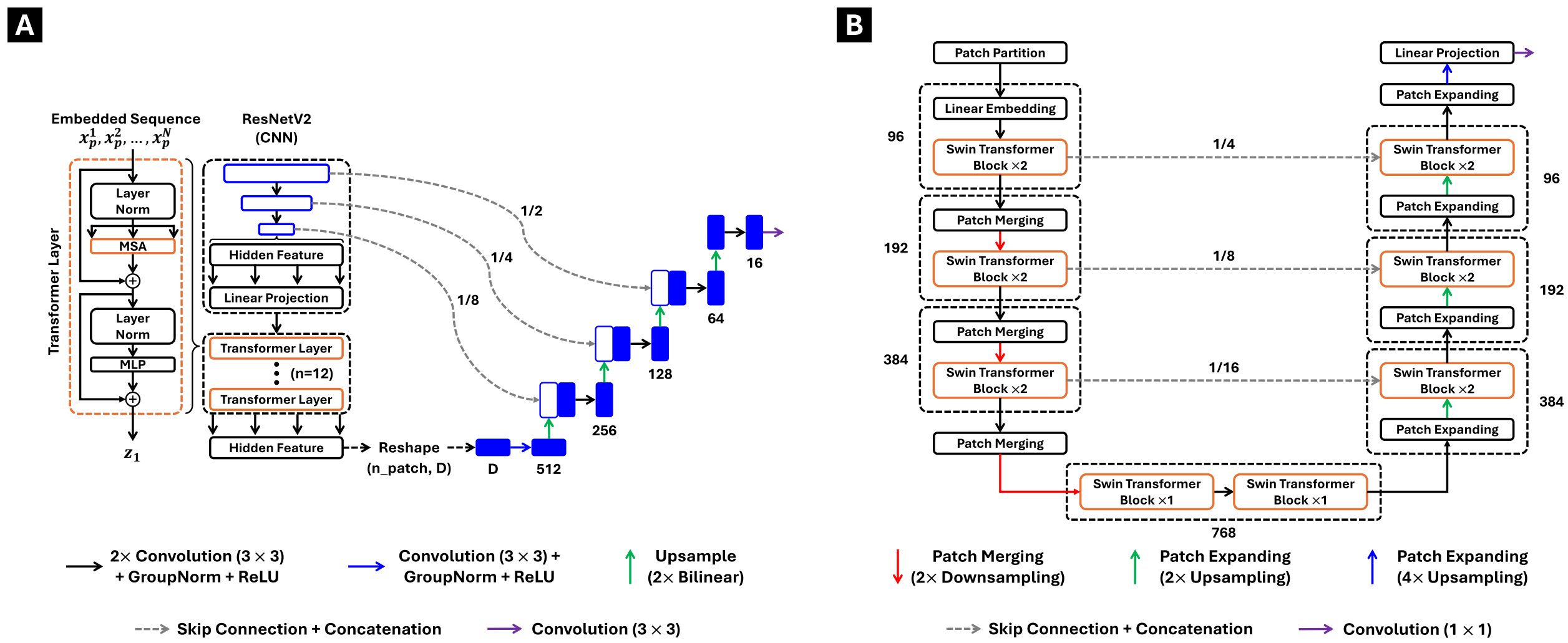

**Fig. 4.** Hybrid convolutional neural network (CNN)-transformer-based and pure transformer-based semantic segmentation architectures implemented to predict binary masks of the vocal folds from two-dimensional (2D) intraglottal particle image velocimetry (PIV) high-speed images. (A) TransUNet and (B) Swin-UNet.

$$z_l' = \mathrm{MSA}\big(\mathrm{LN}(z_{l-1})\big) + z_{l-1} \tag{2}$$

$$z_l = \mathrm{MLP}\big(\mathrm{LN}(z_l')\big) + z_l' \tag{3}$$

Where $\mathrm{LN}(\cdot)$ denotes layer normalization. Self-attention was computed from query (Q), key (K), and value (V) projections as in (4).

$$\mathrm{Attention}(\mathrm{Q}, \mathrm{K}, \mathrm{V}) = \mathrm{softmax}\left(\frac{\mathrm{QK}^{\mathrm{T}}}{\sqrt{d_k}}\right)\mathrm{V} \tag{4}$$

Where $d_k$ is the key dimension. Each layer used 12 attention heads of dimension 64 and an MLP hidden dimension of 3072. Dropout (rate = 0.1) was applied after position-embedding addition, after the attention output projection, and after each MLP linear layer, with no attention-probability dropout.

After a final LN, the token sequence was reshaped to a 32 × 32 map and projected to 512 channels by a 3 × 3 convolution (padding = 1), followed by GN (groups = 8) and a ReLU activation before entering the cascaded upsampler (CUP) decoder. The CUP decoder comprised four 2× bilinear upsampling stages with 256, 128, 64, and 16 channels, each followed by $\mathrm{H}(\cdot)$. In the first three stages, the upsampled features were concatenated with the corresponding encoder skip features before $\mathrm{H}(\cdot)$. A final 3 × 3 convolution produced the single-channel 512 × 512 raw logit map. In the from-scratch pipeline, convolutional weights were initialized using the Kaiming normal method, except the final 3 × 3 segmentation head, which used Xavier normal initialization. Transformer linear weights and position embeddings were drawn from a truncated normal distribution ($\sigma = 0.02$), and biases, where present, were initialized to zero.

Certain modifications were made to the original TransUNet implementation. GN (groups = 8) replaced BN in the CUP decoder, consistent with the CNN-based pipelines, whereas the CNN encoder retained GN (groups = 32). Dropout after the attention output projection followed the original ViT-B/16 formulation [26], whereas the original TransUNet implementation applied no dropout at this position. In addition, the from-scratch pipeline accepted the single-channel grayscale input directly rather than a three-channel image.

The pretrained pipeline replicated the grayscale image across three channels and normalized each channel using the ImageNet means and standard deviations (SDs). It was initialized from the R50+ViT-B/16 ImageNet-21k checkpoint, transferring the ResNetV2 backbone, 1 × 1 token projection, position embeddings, all 12 transformer layers, and the final encoder LN. The CUP decoder and segmentation head were initialized as in the from-scratch pipeline. After the classification-token embedding was discarded, the checkpoint's 14 × 14 position embeddings were bilinearly interpolated to the 32 × 32 grid required for the 512 × 512 input.

### 2.6.4 Swin-UNet

The final benchmarked architecture was the Swin-UNet [29], a symmetric U-shaped, purely transformer-based hierarchical network, as detailed in Fig. 4B. It is built from Swin transformer blocks [39], which follow the layer structure of (2) and (3) but restrict self-attention to local non-overlapping windows. MSA is replaced by window-based MSA (W-MSA) and shifted-window MSA (SW-MSA) in alternating consecutive blocks, enabling cross-window interactions. Within each window, attention incorporates a learned relative-position bias $\mathrm{B}$, which modifies (4) to (5).

$$\mathrm{Attention}(\mathrm{Q}, \mathrm{K}, \mathrm{V}) = \mathrm{softmax}\left(\frac{\mathrm{QK}^{\mathrm{T}}}{\sqrt{d_k}} + \mathrm{B}\right)\mathrm{V} \tag{5}$$

SW-MSA additionally applies an attention mask that restricts interactions across the boundaries introduced by the cyclic window shift. Unlike the TransUNet, the Swin-UNet added no absolute position embedding. Positional information was conveyed solely by $\mathrm{B}$.

The implemented Swin-UNet tokenized the 512 × 512 input using a 4 × 4 convolution (stride = 4) into 96-dimensional patch embeddings, followed by LN, producing a 128 × 128 token grid. The encoder comprised four stages with depths of [2, 2, 2, 2] (two blocks per stage) and 3, 6, 12, and 24 attention heads, respectively, with the fourth stage serving as the bottleneck. All blocks used an 8 × 8 attention window and a GELU-activated MLP with a hidden dimension four times the stage embedding dimension. Patch merging after each of the first three stages halved the spatial resolution and doubled the embedding dimension, giving token grids of $128 \times 128$, $64 \times 64$, $32 \times 32$, and $16 \times 16$ with embedding dimensions of 96, 192, 384, and 768.

The decoder mirrored the encoder, using patch expansion to double the spatial resolution and halve the embedding dimension. Token sequences entering the first three encoder stages were concatenated with decoder features at the corresponding resolutions and linearly projected to the stage dimension. The decoder transformer stages operated at 384, 192, and 96 dimensions. A final 4× patch expansion restored the $512 \times 512$ resolution, after which a $1 \times 1$ convolution produced the single-channel raw logit map. All normalization in the network was performed using LN. Standard dropout was disabled (rate = 0.0), whereas stochastic depth increased linearly from 0 to 0.1 across the eight encoder blocks, with decoder blocks using the rates of their mirrored encoder stages. In the from-scratch pipeline, relative-position-bias tables and linear and convolutional weights were drawn from a truncated normal distribution ($\sigma = 0.02$), except the $1 \times 1$ segmentation head, which used Xavier normal initialization. Biases, where present, were initialized to zero.

This study's Swin-UNet implementation differed from the original by using an $8 \times 8$ attention window instead of the $7 \times 7$ window so that the windows evenly divided the token grid at every stage without padding. The maximum stochastic-depth rate was 0.1, matching the configuration of the pretrained checkpoint, rather than 0.2 of the original Swin-UNet. Additionally, the from-scratch pipeline accepted the single-channel grayscale input directly rather than a three-channel image.

The pretrained pipeline replicated the grayscale image across three channels and applied the same ImageNet normalization as the pretrained TransUNet. It was initialized from a Swin-Tiny checkpoint trained on ImageNet-22k and fine-tuned on ImageNet-1k, transferring the patch embedding, all four encoder stages including patch merging, and the final encoder LN. Source stages 1–3 also initialized the corresponding decoder transformer stages, whereas source stage 4 initialized only the encoder bottleneck. Because the source checkpoint used stage depths of [2, 2, 6, 2], only the first two blocks of its third stage were transferred. Patch expansion layers, skip-projection layers, the decoder LN, the final 4× expansion, and the segmentation head were initialized as in the from-scratch pipeline. The source $13 \times 13$ relative-position-bias tables (window 7) were bicubically interpolated to the $15 \times 15$ tables required for window 8, and the relative-position indices and shifted-window masks were recomputed.

## 2.7 Standardized training protocol

All six semantic segmentation pipelines were trained using a standardized three-stage protocol, whose shared settings and hyperparameters are summarized in Table 1. All models were implemented in PyTorch (version 2.11.0) and trained on an NVIDIA RTX PRO 6000 Blackwell Server Edition GPU with CUDA 12.8 and cuDNN 9.19.0.

**Table 1.** Shared training settings and hyperparameters defining the standardized three-stage training protocol for the six benchmarked semantic segmentation pipelines

| Setting | | Standardized Protocol |
|---|---|---|
| 1. | Input image resolution | $512 \times 512$ |
| 2. | Data augmentation | None |
| 3. | Loss function | Unweighted binary cross-entropy with logits; mean reduction |
| 4. | Optimizer | AdamW; $\beta_1 = 0.9$, $\beta_2 = 0.999$, $\varepsilon = 1\times10^{-8}$, weight decay = $1\times10^{-4}$ |
| 5. | Batch size | 16 |
| 6. | Candidate peak learning rate grid | $\{1\times10^{-2}, 1\times10^{-3}, 1\times10^{-4}, 1\times10^{-5}\}$ |
| 7. | Learning rate warmup | Batch-wise linear from $0.01\times\eta_{\text{peak}}$ to $\eta_{\text{peak}}$ over the first 2 epochs; applied in every training run |
| 8. | Learning rate schedule (Stages 1–2) | Post-warmup: reduce learning rate on plateau of validation loss; factor = 0.5, patience = 5 epochs, floor = $1\times10^{-7}$ |
| 9. | Learning rate schedule (Stage 3) | Post-warmup: cosine annealing from $\eta_{\text{peak}}$ to $1\times10^{-7}$ over the remaining predetermined training epochs |
| 10. | Maximum training duration (Stages 1–2) | 40 epochs |
| 11. | Early stopping (Stages 1–2) | Validation loss; patience = 10 epochs |
| 12. | Mixed precision | Automatic mixed precision with dynamic loss scaling |
| 13. | Reproducibility | Stages 1–2: seed = 26. Stage 3: seeds {6, 12, 24, 48, 96}, identical across pipelines; deterministic cuDNN execution with autotuning disabled |

In Stage 1, the peak learning rate (LR) of each pipeline was selected. The frozen training partition was divided into five folds, so that all images from a given trial remained within the same fold. Fold 1 was fixed *a priori* as the validation fold, and the other four were combined into a single training set. Each candidate peak LR in Table 1 was evaluated on this split using a newly instantiated model initialized with seed = 26. Each run ended at the early-stopping criterion or the maximum training duration defined in Table 1. The checkpoint with the lowest validation loss, which could precede the final epoch, was then restored for evaluation. The LR yielding the highest mean per-image Dice similarity coefficient (DSC) on the validation fold was selected for Stages 2 and 3.

In Stage 2, five-fold cross-validation (CV) was performed over the same five folds using the selected peak LR. For each fold, a new model (seed = 26) was trained on the other four folds and evaluated on the validation fold after its lowest validation-loss checkpoint was restored, as in Stage 1. The epoch corresponding to this checkpoint was designated the best epoch. Importantly, all five folds were required to complete. The mean validation DSC across the five folds was used to rank the pipelines for the prespecified top-two comparison, as described in Section 2.8. The mean and sample SD of the five fold-level validation DSC, boundary F1 score (BF1), and 95th percentile Hausdorff distance (HD95) were also recorded. The Stage 3 training duration was derived by taking the arithmetic mean of the five best epochs, rounded to the nearest integer, with exact half-integer values rounded upward. This follows the established practice of using early stopping to determine the training duration and then retraining on all available training data for that duration [40].

In Stage 3, each pipeline was trained on the complete training partition, without validation or early stopping, using the selected Stage 1 peak LR and the calculated Stage 2 training duration. Fixed-duration training is widely used in biomedical image segmentation [41] and was required here because the Stage 3 cosine schedule depends on the total number of training epochs. Training was repeated with each of the five Stage 3 seeds as detailed in Table 1. Each training run initialized a new model and a new seeded training data loader, with the final weights being saved for held-out test evaluation. The held-out test set was not used in any stage of model development, and its results did not influence the model weights, peak LR, training duration, seeds, or LR schedule. For qualitative visualization, the Stage 3 seed with the median held-out test DSC for each pipeline was chosen as the representative seed.

Two training settings applied only to subsets of the pipelines. In the pretrained TransUNet and Swin-UNet pipelines, transferred parameters remained trainable throughout optimization, with no layer freezing. In all four hybrid CNN-transformer- and transformer-based pipelines, global L2-norm gradient clipping (maximum norm = 1.0) was applied after the mixed-precision gradients were unscaled. No clipping was applied to the UNet or UNet++ pipelines.

### 2.8 Evaluation metrics and statistical testing

Segmentation performance on the held-out test set was evaluated using three primary metrics encompassing overall regional overlap and boundary fidelity: DSC, BF1, and HD95 [42,43]. Predicted masks were obtained by applying the sigmoid function to the raw model outputs and assigning pixels with probability $> 0.5$ to the foreground. Let $\mathrm{P}$ and $\mathrm{G}$ denote the foreground pixel sets of the predicted and ground-truth masks, respectively, and let $\partial\mathrm{P}$ and $\partial\mathrm{G}$ denote their one-pixel inner boundaries, obtained by subtracting a single binary erosion from each mask with a 4-connected cross-shaped structuring element. The Euclidean distance from a pixel $x$ to the nearest pixel in a set $\mathrm{S}$ was defined as $d(x,\mathrm{S}) = \min_{s \in \mathrm{S}} \|x - s\|_2$, measured in pixels at the $512 \times 512$ output resolution. The primary metrics were defined as in (6)–(8).

$$\mathrm{DSC} = \frac{2|\mathrm{P} \cap \mathrm{G}|}{|\mathrm{P}| + |\mathrm{G}|} \tag{6}$$

$$\mathrm{BP} = \frac{|\{p \in \partial\mathrm{P} : d(p, \partial\mathrm{G}) \leq \tau\}|}{|\partial\mathrm{P}|} \tag{7}$$

$$\mathrm{BR} = \frac{|\{g \in \partial \mathrm{G}: d(g, \partial \mathrm{P}) \leq \tau\}|}{|\partial \mathrm{G}|}$$

$$\mathrm{BF1} = \frac{2 \times \mathrm{BP} \times \mathrm{BR}}{\mathrm{BP} + \mathrm{BR}}$$

$$\mathrm{HD95} = \max\{\mathrm{Q}_{95}(\{d(p, \partial \mathrm{G}): p \in \partial \mathrm{P}\}), \mathrm{Q}_{95}(\{d(g, \partial \mathrm{P}): g \in \partial \mathrm{G}\})\} \quad (8)$$

Where BP and BR denote boundary precision and boundary recall, respectively, $\tau = 2$ pixels is the boundary tolerance, and $\mathrm{Q}_{95}(\cdot)$ denotes the 95th percentile. The 2-pixel tolerance was selected to emphasize high-fidelity vocal fold delineation. DSC and BF1 range from 0 to 1, with higher values indicating better agreement, whereas HD95 is expressed in pixels, with lower values indicating better agreement. HD95 is undefined when exactly one of the two masks is empty. Such values were treated as non-finite, excluded from all descriptive summaries, and counted separately.

Overall pipeline performance was reported as the mean and sample SD of the five seed-level averages, each computed from the per-image values of one Stage 3 seed. Image-level variability in DSC and BF1 was summarized by the sample SD of the per-image values within each seed. The arithmetic mean of these five SDs was reported, together with their range across the seeds. Because per-image HD95 values can be strongly right-skewed, HD95 variability was instead summarized by the median, the interquartile range (IQR), and the 90th percentile. Large boundary errors were quantified as the proportion of images with HD95 $>$ 50 pixels, approximately 10% of the image width. These HD95 statistics were computed once from the per-image values of all five seeds combined (5 seeds $\times$ 3,600 held-out test set images) to describe the error distribution across all trained models of each pipeline.

Formal pipeline comparisons were performed at the trial level to avoid treating the 72 correlated phase-resolved images within a phonation trial as independent observations [44]. For a contrast between pipelines $\mathrm{A}$ and $\mathrm{B}$, each held-out test trial contributed one paired difference $(\mathrm{A} - \mathrm{B})$. For DSC and BF1, per-image values were averaged within each trial, separately for each pipeline and seed. These trial-level values were then averaged across the five seeds, yielding one value per pipeline and trial, from which the paired difference was calculated. For HD95, the paired difference was instead calculated for each image, using only images with finite values in both pipelines. These per-image differences were then averaged within each trial, separately for each seed, and then across the five seeds. Non-finite values were neither capped nor replaced, and their number was reported for each contrast. Both procedures resulted in one paired difference per trial, with positive DSC and BF1 differences and negative HD95 differences favoring $\mathrm{A}$.

Three contrasts were prespecified: (1) the two pipelines ranked highest by mean Stage 2 CV DSC, identified from the training partition alone, with the higher-ranked pipeline as $\mathrm{A}$; (2) TransUNet trained from scratch $(\mathrm{A})$ versus pretrained $(\mathrm{B})$; and (3) Swin-UNet trained from scratch $(\mathrm{A})$ versus pretrained $(\mathrm{B})$. For each primary metric and contrast, the paired trial differences were tested against zero using a two-sided Wilcoxon signed-rank test [45]. Holm-Bonferroni correction (hereafter Holm) was applied across the three contrasts within each metric [46], and a Holm-adjusted $p \leq 0.05$ was considered statistically significant. The primary effect estimate was the median paired trial difference, with a 95% percentile-bootstrap confidence interval (CI) obtained from 10,000 resamples of trials with replacement. Similar statistical testing approaches were used in prior benchmarking studies [33,47]. To assess sensitivity to the training seed, the median paired trial difference was also calculated from each seed's models separately, using the same trials, and the range of these five medians was reported. These seed-specific medians were descriptive only and were not tested, because all five seeds were evaluated on the same test trials.

Exploratory *post hoc* contrasts compared CNN-based pipelines to the hybrid CNN-transformer- and the pure transformer-based pipelines. Specifically, the UNet and UNet++ were each compared to the TransUNet and Swin-UNet trained from scratch, using the same trial-level procedure with the CNN-based pipeline as $\mathrm{A}$. Holm correction was applied across these four contrasts within each metric, separately from the prespecified contrasts.

## 3. Results

### 3.1 Learning rate selection and cross-validation segmentation performance

The peak LR and Stage 3 training duration of each semantic segmentation pipeline were determined in Stages 1 and 2 using only the training partition. Table 2 summarizes these values together with the number of trainable parameters. The selected peak LR of the pretrained TransUNet ($1 \times 10^{-5}$) was the lowest value in the candidate grid, whereas the highest candidate ($1 \times 10^{-2}$) was not selected for any pipeline. The from-scratch and pretrained variants of the TransUNet and Swin-UNet differed in parameter count due to the input layer. The from-scratch pipelines accepted single-channel inputs while the pretrained pipelines repeated the grayscale input across three channels.

**Table 2.** Trainable parameters, selected peak learning rate, and Stage 3 training duration for the six benchmarked semantic segmentation pipelines

| Segmentation Pipeline | Trainable Parameters | Peak Learning Rate | Stage 3 Training Duration (epochs) |
|---|---|---|---|
| UNet | 31,042,369 | $1 \times 10^{-4}$ | 11 |
| UNet++ | 36,171,393 | $1 \times 10^{-3}$ | 10 |
| TransUNet (*scratch*) | 105,905,553 | $1 \times 10^{-4}$ | 6 |
| TransUNet (*pretrained*) | 105,911,825 | $1 \times 10^{-5}$ | 9 |
| Swin-UNet (*scratch*) | 27,172,453 | $1 \times 10^{-4}$ | 16 |
| Swin-UNet (*pretrained*) | 27,175,525 | $1 \times 10^{-3}$ | 10 |

Table 3 reports, for each pipeline, the mean and sample SD of the five fold-level validation DSC, BF1, and HD95 values obtained in Stage 2 with the selected peak LR. All five folds were completed for every pipeline, and the corresponding loss trajectories are shown in Fig. 5. Mean CV DSC spanned a narrow range across pipelines (0.9721–0.9756), and mean BF1 ranged from 0.6337 to 0.6680. A larger difference in mean CV HD95 was observed across pipelines, ranging from 40.30 to 23.65 pixels. The two CNN-based pipelines had the highest HD95, at 40.30 pixels for the UNet and 37.85 pixels for the UNet++. Fold-to-fold SDs did not exceed 0.0023 for DSC or 0.0070 for BF1. For HD95, they were smaller for the pretrained TransUNet and pretrained Swin-UNet (2.07 and 3.85 pixels) than for the remaining four pipelines (5.13–7.29 pixels). The same two pipelines ranked first and second, respectively, on all three metrics. Because the prespecified ranking was based on mean CV DSC (0.9756 and 0.9752), they formed the first prespecified contrast, with the pretrained TransUNet as $\mathrm{A}$.

**Table 3.** Five-fold CV DSC, BF1, and HD95 (mean ± sample SD across folds) for the six benchmarked semantic segmentation pipelines

| Segmentation Pipeline | DSC | BF1 | HD95 (pixels) |
|---|---|---|---|
| UNet | 0.9728 ± 0.0017 | 0.6449 ± 0.0053 | 40.30 ± 5.15 |
| UNet++ | 0.9734 ± 0.0019 | 0.6506 ± 0.0051 | 37.85 ± 5.13 |
| TransUNet (*scratch*) | 0.9721 ± 0.0023 | 0.6524 ± 0.0070 | 34.47 ± 7.29 |
| TransUNet (*pretrained*) | **0.9756 ± 0.0012** | **0.6680 ± 0.0034** | **23.65 ± 2.07** |
| Swin-UNet (*scratch*) | 0.9727 ± 0.0015 | 0.6337 ± 0.0061 | 35.37 ± 5.33 |
| Swin-UNet (*pretrained*) | **0.9752 ± 0.0015** | **0.6579 ± 0.0062** | **25.73 ± 3.85** |

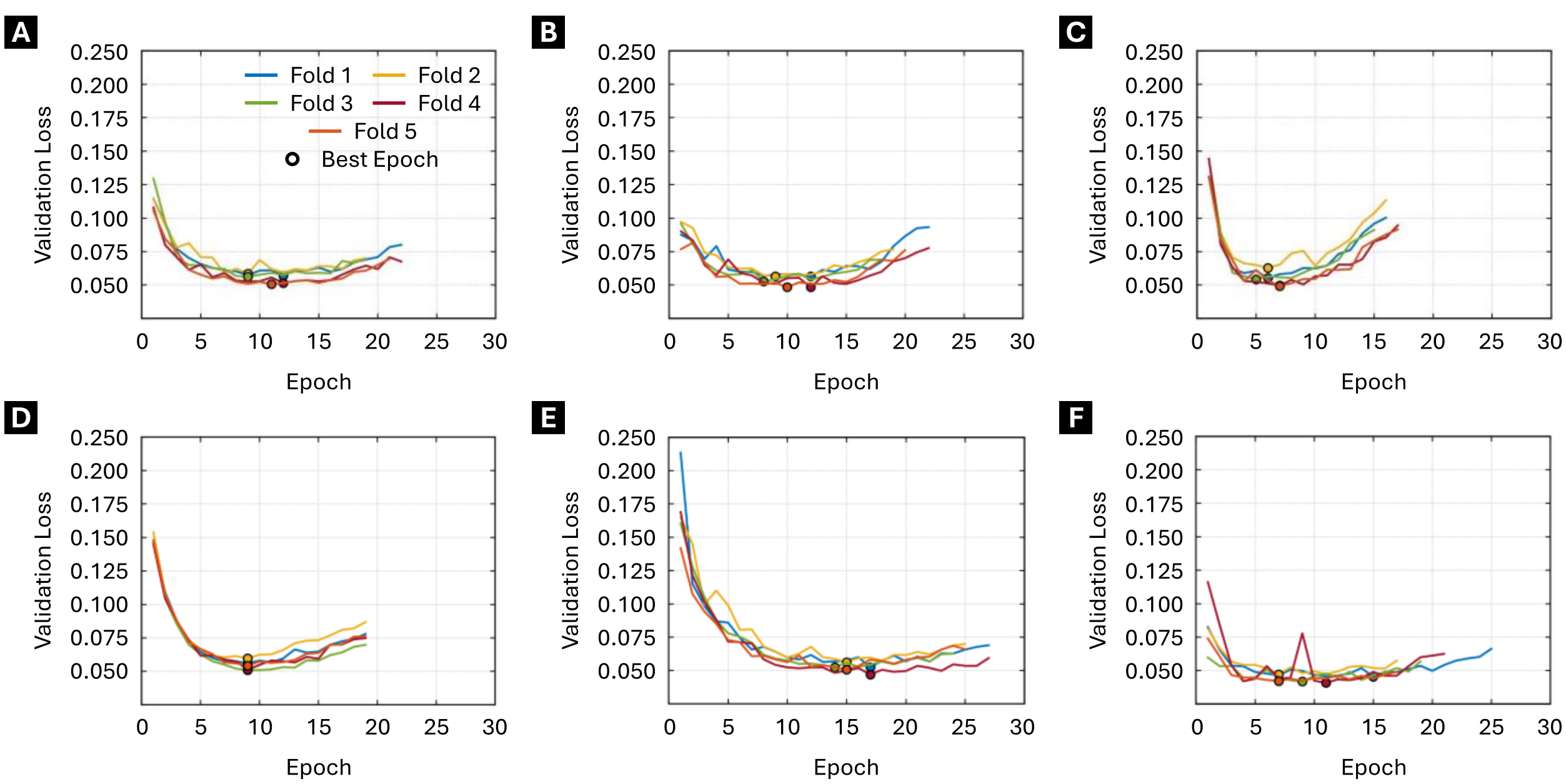

**Fig. 5.** Stage 2 five-fold cross-validation (CV) loss trajectories for the six benchmarked semantic segmentation pipelines, with filled markers indicating the saved model checkpoint at the epoch corresponding to the lowest validation loss for each fold. (A) UNet, (B) UNet++, (C) TransUNet (*scratch*), (D) TransUNet (*pretrained*), (E) Swin-UNet (*scratch*), (F) Swin-UNet (*pretrained*).

## 3.2 Held-out test segmentation performance

In Stage 3, each pipeline was trained on the full frozen training partition over five seeds, using its respective peak LR and training duration. The resulting 30 models (six pipelines × five seeds) were then evaluated on the held-out test set. Table 4 reports the mean and sample SD of the five seed-level averages of DSC, BF1, and HD95 for each segmentation pipeline. No ground-truth or predicted mask was empty. Therefore, all per-image HD95 values were finite. Mean DSC again spanned a narrow range across pipelines (0.9734–0.9764), and mean BF1 ranged from 0.6425 to 0.6679. Mean HD95 distinguished the pipelines more clearly, ranging from 34.36 to 23.61 pixels. As in Stage 2, the TransUNet and Swin-UNet pipelines, regardless of pretraining, had lower mean HD95 than the UNet and UNet++. The pretrained Swin-UNet and the pretrained TransUNet ranked first and second, respectively, on all three metrics, reversing their Stage 2 order. Seed-to-seed sample SDs did not exceed 0.0004 for DSC or 0.0020 for BF1. For HD95, they were larger for the UNet and UNet++ (2.05 and 2.68 pixels) than for the TransUNet and Swin-UNet pipelines (0.30–0.97 pixels).

**Table 4.** Held-out test DSC, BF1, and HD95 (mean $\pm$ sample SD across seeds) for the six benchmarked semantic segmentation pipelines

| Segmentation Pipeline | DSC | BF1 | HD95 (pixels) |
|---|---|---|---|
| UNet | 0.9743 ± 0.0001 | 0.6495 ± 0.0009 | 34.36 ± 2.05 |
| UNet++ | 0.9749 ± 0.0003 | 0.6552 ± 0.0016 | 33.60 ± 2.68 |
| TransUNet (*scratch*) | 0.9740 ± 0.0003 | 0.6594 ± 0.0013 | 26.42 ± 0.97 |
| TransUNet (*pretrained*) | **0.9756 ± 0.0004** | **0.6670 ± 0.0020** | **25.02 ± 0.30** |
| Swin-UNet (*scratch*) | 0.9734 ± 0.0001 | 0.6425 ± 0.0009 | 30.83 ± 0.54 |
| Swin-UNet (*pretrained*) | **0.9764 ± 0.0001** | **0.6679 ± 0.0009** | **23.61 ± 0.60** |

Table 5 summarizes how performance varied across individual held-out test images. For DSC and BF1, it reports the mean of the five within-seed per-image SDs, with their range across seeds. For HD95, it reports instead the median, IQR, and $90^{th}$ percentile of the 18,000 per-image values of each pipeline (five seeds × 3,600 images), together with the proportion of images with HD95 $>$ 50 pixels. For every pipeline, the per-image SDs exceeded the corresponding seed-to-seed SDs in Table 4 by more than an order of magnitude (DSC: 0.0123–0.0142 vs $\leq$ 0.0004; BF1: 0.0589–0.0669 vs $\leq$ 0.0020). The pretrained Swin-UNet and the pretrained TransUNet had the lowest per-image DSC SDs (0.0123 and 0.0129), whereas the from-scratch TransUNet and the from-scratch Swin-UNet had the highest (0.0141 and 0.0142). For BF1, the UNet and the from-scratch Swin-UNet had the lowest per-image SDs (0.0589 and 0.0594). The highest per-image BF1 SDs belonged to the pretrained Swin-UNet and the pretrained TransUNet (0.0665 and 0.0669), the two pipelines with the highest mean BF1 in Table 4.

The pretrained Swin-UNet and the pretrained TransUNet had the lowest median HD95 (15.2 and 15.8 pixels), the narrowest IQRs (11.0–23.6 and 11.0–24.4 pixels), and the lowest 90$^{th}$ percentiles (52.0 and 52.8 pixels). They also had the smallest proportions of images with HD95 > 50 pixels (10.3% and 10.5%), less than half of those of the UNet++ (21.3%) and UNet (22.0%). The UNet had the highest median HD95 (23.5 pixels) and the from-scratch Swin-UNet had the highest 90$^{th}$ percentile (80.4 pixels), close to those of the UNet and UNet++ (80.0 and 79.8 pixels).

**Table 5.** Held-out test per-image variability in DSC, BF1, and HD95 for the six benchmarked semantic segmentation pipelines

| Segmentation Pipeline | DSC per-image SD, mean (range across seeds) | BF1 per-image SD, mean (range across seeds) | HD95 median [IQR] (pixels) | HD95 90$^{th}$ percentile (pixels) | Images with HD95 > 50 pixels (%) |
|---|---|---|---|---|---|
| UNet | 0.0133 (0.0131–0.0135) | **0.0589 (0.0588–0.0590)** | 23.5 [14.2–45.1] | 80.0 | 22.0 |
| UNet++ | 0.0130 (0.0128–0.0134) | 0.0596 (0.0586–0.0603) | 21.9 [13.6–43.4] | 79.8 | 21.3 |
| TransUNet (*scratch*) | 0.0141 (0.0138–0.0147) | 0.0612 (0.0603–0.0618) | 17.0 [12.2–27.6] | 57.6 | 11.6 |
| TransUNet (*pretrained*) | **0.0129 (0.0125–0.0133)** | 0.0669 (0.0663–0.0673) | **15.8 [11.0–24.4]** | **52.8** | **10.5** |
| Swin-UNet (*scratch*) | 0.0142 (0.0139–0.0147) | **0.0594 (0.0590–0.0599)** | 19.4 [13.4–33.9] | 80.4 | 15.8 |
| Swin-UNet (*pretrained*) | **0.0123 (0.0122–0.0124)** | 0.0665 (0.0658–0.0674) | **15.2 [11.0–23.6]** | **52.0** | **10.3** |

### 3.3 Statistical testing

Formal pipeline comparisons were done at the trial level, with each of the 50 held-out test trials contributing one paired difference per contrast and metric. Because all per-image HD95 values were finite, every HD95 contrast used all 3,600 images present in the 50 trials. Table 6 reports the three prespecified contrasts as the median paired trial difference $(\mathrm{A} - \mathrm{B})$ with its 95% CI. Positive DSC and BF1 differences and negative HD95 differences favor pipeline $\mathrm{A}$. All $p$ values are Holm-adjusted, and differences with $p \leq 0.05$ are described as statistically significant. Seven of the nine tests were statistically significant, including all six tests pertaining to the two pretraining contrasts and the DSC test of the top-two contrast. In every test, the 95% CI excluded zero when the difference was significant and included zero when it was not.

In the top-two contrast, the pretrained Swin-UNet (B) had a significantly higher DSC than the pretrained TransUNet (A) (−0.0010 [−0.0015, −0.0004]; $p = 0.001$), and this difference favored the pretrained Swin-UNet across all five seeds. The BF1 and HD95 differences were not statistically significant (BF1: −0.0004 [−0.0036, +0.0018], $p = 0.619$; HD95: +0.82 [−0.26, +1.70] pixels, $p = 0.121$). For both metrics, the 95% CI included zero, and the seed-specific medians favored the pretrained TransUNet for some seeds and the pretrained Swin-UNet for others.

In both pretraining contrasts, the pretrained pipeline (B) had a significantly higher DSC and BF1 and a significantly lower HD95 than its from-scratch counterpart (A), and all five seed-specific medians favored the pretrained pipeline in every case. For the TransUNet, the differences were −0.0010 [−0.0020, −0.0002] for DSC ($p < 0.001$), −0.0063 [−0.0104, −0.0028] for BF1 ($p < 0.001$), and +1.31 [+0.42, +2.32] pixels for HD95 ($p = 0.044$). This HD95 difference had the largest $p$ value of the seven significant results. For the Swin-UNet, the differences were −0.0024 [−0.0033, −0.0017] for DSC, −0.0250 [−0.0293, −0.0213] for BF1, and +5.11 [+4.57, +6.64] pixels for HD95 (all $p < 0.001$). These were larger than the corresponding TransUNet differences for every metric, by a factor of about four for BF1 and HD95.

**Table 6.** Prespecified contrasts of held-out test DSC, BF1, and HD95

| **Contrast (A vs B)** | **Metric** | **Median paired trial difference, A − B [95% CI]** | **Holm-adjusted $p$** | **Seed-specific median range** |
|---|---|---|---|---|
| TransUNet (*pretrained*) vs Swin-UNet (*pretrained*) | DSC | −0.0010 [−0.0015, −0.0004] | 0.001 | −0.0015 to −0.0006 |
| | BF1 | −0.0004 [−0.0036, +0.0018] | 0.619 | −0.0033 to +0.0024 |
| | HD95 (pixels) | +0.82 [−0.26, +1.70] | 0.121 | −0.24 to +1.37 |
| TransUNet (*scratch*) vs TransUNet (*pretrained*) | DSC | −0.0010 [−0.0020, −0.0002] | $< 0.001$ | −0.0015 to −0.0007 |
| | BF1 | −0.0063 [−0.0104, −0.0028] | $< 0.001$ | −0.0082 to −0.0059 |
| | HD95 (pixels) | +1.31 [+0.42, +2.32] | 0.044 | +0.71 to +2.58 |
| Swin-UNet (*scratch*) vs Swin-UNet (*pretrained*) | DSC | −0.0024 [−0.0033, −0.0017] | $< 0.001$ | −0.0027 to −0.0024 |
| | BF1 | −0.0250 [−0.0293, −0.0213] | $< 0.001$ | −0.0267 to −0.0244 |
| | HD95 (pixels) | +5.11 [+4.57, +6.64] | $< 0.001$ | +4.57 to +6.61 |

Table 7 reports the four exploratory *post hoc* contrasts, which compared each CNN-based pipeline (A) to the hybrid CNN-transformer- and the pure transformer-based pipelines trained from scratch ($\mathrm{B}$). Because these contrasts were not prespecified, their $p$ values were Holm-adjusted within this family of four contrasts, separately from those in Table 6. Positive DSC and BF1 differences and negative HD95 differences therefore favor the CNN-based pipeline. Nine of the 12 tests were statistically significant. The 95% CI agreed with the significance decision in 10 tests. The exceptions were the BF1 and HD95 tests of the UNet++ versus the from-scratch TransUNet, which were significant although their CIs included zero. This can occur because the Wilcoxon test uses the signs and ranks of all 50 paired differences, whereas the CI describes only their median.

Compared with both CNN-based pipelines, TransUNet trained from scratch ($\mathrm{B}$) had a significantly higher BF1 and a lower HD95. Against the UNet, the differences were −0.0100 [−0.0116, −0.0038] for BF1 ($p < 0.001$) and +6.30 [+0.23, +13.22] pixels for HD95 ($p = 0.001$), and all five seed-specific medians favored the from-scratch TransUNet for both metrics. Against the UNet++, the differences were −0.0020 [−0.0049, +0.0007] for BF1 ($p = 0.039$) and +4.71 [−0.14, +9.17] pixels for HD95 ($p < 0.001$). All five seed-specific medians favored the from-scratch TransUNet for HD95 but not for BF1. The DSC differences did not exceed 0.0009. The UNet++ had a significantly higher DSC than the from-scratch TransUNet (+0.00085 [+0.00003, +0.00131]; $p =$ 0.024), whereas the UNet difference was not statistically significant (+0.0005 [−0.0003, +0.0009]; $p = 0.534$).

Compared with the Swin-UNet trained from scratch ($\mathrm{B}$), both CNN-based pipelines had a significantly higher DSC and BF1, and all five seed-specific medians favored the CNN-based pipelines for both metrics. For the UNet, the differences were +0.0008 [+0.0002, +0.0015] for DSC ($p = 0.024$) and +0.0104 [+0.0055, +0.0129] for BF1 ($p = 0.003$). For the UNet++, they were +0.0014 [+0.0008, +0.0021] for DSC and +0.0151 [+0.0115, +0.0190] for BF1 (both $p < 0.001$). The HD95 differences were not statistically significant (UNet: +3.35 [−2.18, +6.68] pixels; UNet++: +1.03 [−1.64, +4.57] pixels; both $p = 0.142$).

**Table 7.** Exploratory *post hoc* contrasts of held-out test DSC, BF1, and HD95

| **Contrast ($\mathrm{A}$ vs $\mathrm{B}$)** | **Metric** | **Median paired trial difference, $\mathrm{A} - \mathrm{B}$ [95% CI]** | **Holm-adjusted $p$** | **Seed-specific median range** |
|---|---|---|---|---|
| UNet vs TransUNet (*scratch*) | DSC | +0.0005 [−0.0003, +0.0009] | 0.534 | −0.0002 to +0.0004 |
| | BF1 | −0.0100 [−0.0116, −0.0038] | $< 0.001$ | −0.0119 to −0.0071 |
| | HD95 (pixels) | +6.30 [+0.23, +13.22] | 0.001 | +4.35 to +8.34 |
| UNet vs Swin-UNet (*scratch*) | DSC | +0.0008 [+0.0002, +0.0015] | 0.024 | +0.0006 to +0.0010 |
| | BF1 | +0.0104 [+0.0055, +0.0129] | 0.003 | +0.0058 to +0.0116 |
| | HD95 (pixels) | +3.35 [−2.18, +6.68] | 0.142 | +0.84 to +4.36 |
| UNet++ vs TransUNet (*scratch*) | DSC | +0.00085 [+0.00003, +0.00131] | 0.024 | +0.0002 to +0.0018 |
| | BF1 | −0.0020 [−0.0049, +0.0007] | 0.039 | −0.0056 to +0.0002 |
| | HD95 (pixels) | +4.71 [−0.14, +9.17] | < 0.001 | +1.37 to +9.44 |
| UNet++ vs Swin-UNet (*scratch*) | DSC | +0.0014 [+0.0008, +0.0021] | < 0.001 | +0.0011 to +0.0016 |
| | BF1 | +0.0151 [+0.0115, +0.0190] | < 0.001 | +0.0128 to +0.0169 |
| | HD95 (pixels) | +1.03 [−1.64, +4.57] | 0.142 | −0.63 to +6.65 |

### 3.4 Qualitative segmentation comparison

Qualitative visualizations of segmentation performance were generated for select test trials, using the Stage 3 seed with the median held-out test DSC as the representative seed for each pipeline. The best and worst cases were identified based on the worst per-image HD95 of each trial, defined as the largest HD95 among the 72 phase-resolved images across all six pipelines. The best case was trial 42, whose worst per-image HD95 was 42.19 pixels, and it was the only test trial in which no per-image HD95 value was > 50 pixels in any pipeline. Fig. 6 illustrates the overlay for trial 42 comparing the predicted mask to the ground-truth mask for each pipeline during the closed, early opening, mid-opening, fully open, and closing phases of the vibratory cycle. The images corresponding to these phases were chosen following visual inspection, and each image is labeled with its respective HD95 value.

Overall, the predicted and ground-truth masks were closely aligned in all pipelines. In the closed phase, the ground-truth mask contained a narrow slit at the midline where the vocal folds coapt. The from-scratch TransUNet and the from-scratch Swin-UNet segmented over the entire slit (false positive; red), whereas the other pipelines mildly undersegmented its upper part (false negative; blue) and segmented over its lower part, which was longest in the UNet. The UNet, from-scratch TransUNet, and from-scratch Swin-UNet had the highest HD95 values in this column (9.85–16.03 pixels). During early and mid-opening, undersegmentation appeared along the top of both vocal folds and at the midline when they separated. In the fully open and closing phases, the predicted masks of all pipelines closely matched the ground truth (5.00–11.00 pixels), apart from mild undersegmentation at the bottom of the opening when fully open. Small areas of oversegmentation also remained at closing, at the bottom of the opening for the UNet++ and along its lower left edge for the from-scratch Swin-UNet. The pretrained Swin-UNet had the lowest HD95 at closing (5.00 pixels) and shared the lowest value in the fully open phase (7.00 pixels) with the pretrained TransUNet.

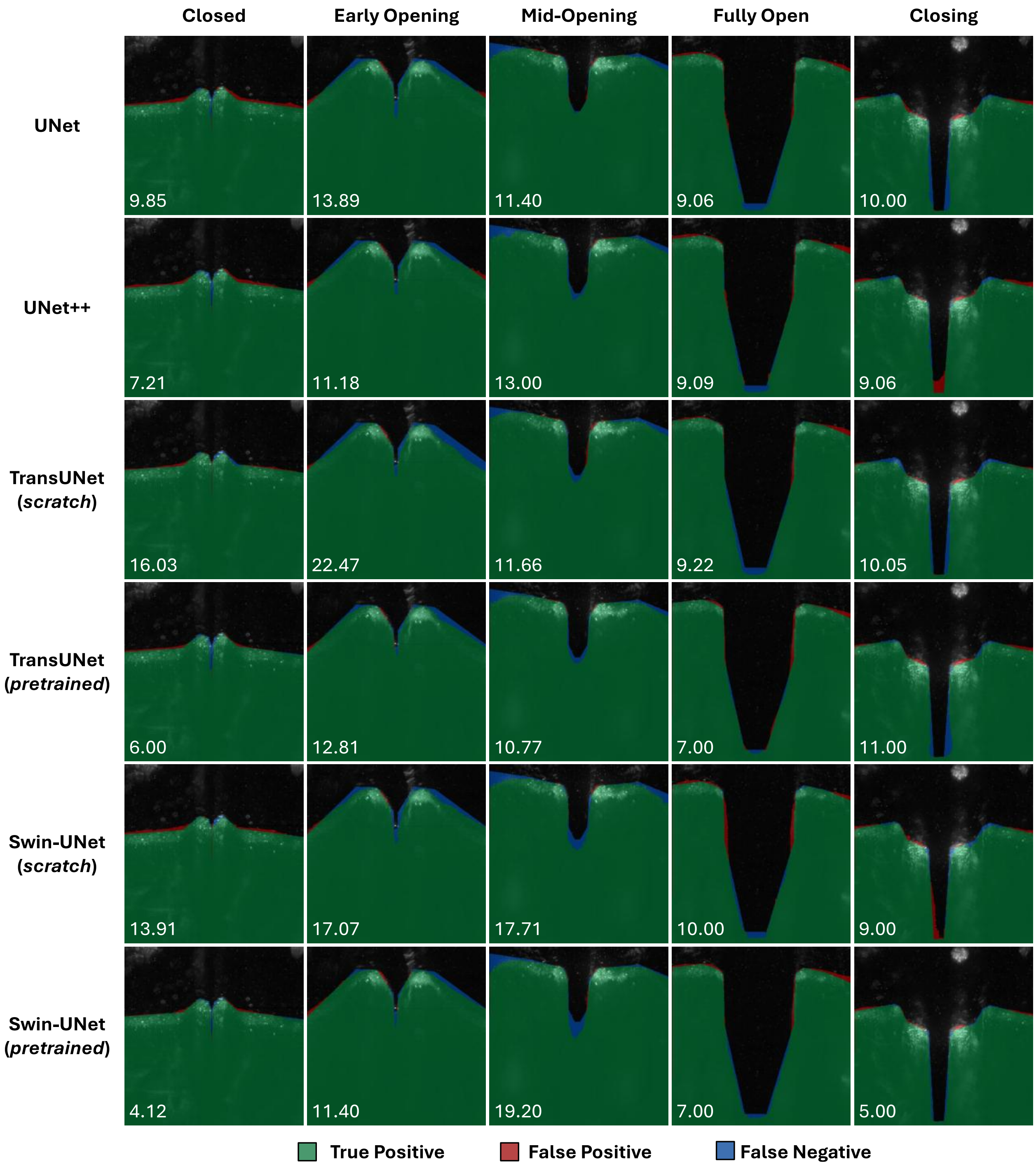

**Fig. 6.** Predicted masks overlaid on the ground-truth masks for the six benchmarked semantic segmentation pipelines in the best case (held-out test trial 42), using each pipeline's representative seed. The numerical value in each image panel denotes its respective HD95 magnitude (pixels).

The worst case was trial 130, whose worst per-image HD95 was 167.02 pixels, the largest among all held-out test trials. At least one pipeline had an HD95 value $> 50$ pixels in 46 of its 72 phase-resolved images. Fig. 7 displays the overlay comparing the predicted and ground-truth masks for trial 130. Similar to trial 42, a narrow slit was visible at the midline of the ground-truth mask during the closed phase. In contrast to the best case, however, all six pipelines oversegmented the entire slit, and their HD95 values were similar (43.00–46.45 pixels). During early opening, the slit in the ground-truth mask extended all the way down the vocal folds. The from-scratch Swin-UNet oversegmented nearly the entire length of this slit (78.70 pixels), while the other pipelines oversegmented it to a lesser degree and left a portion of it open. Additionally, all pipelines showed oversegmentation at the top of the vocal folds and the pretrained TransUNet mildly undersegmented the top corner of the left vocal fold.

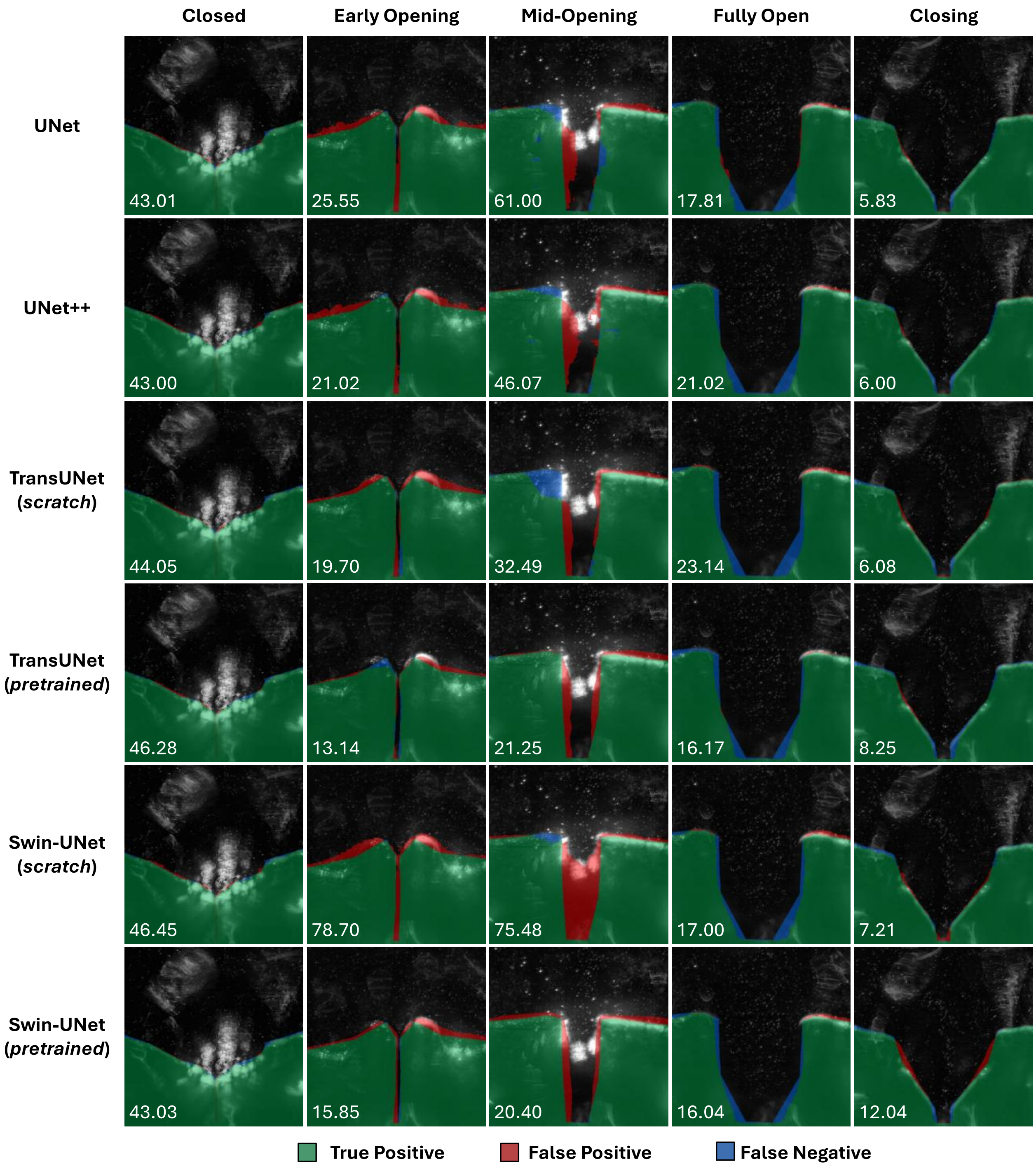

**Fig. 7.** Predicted masks overlaid on the ground-truth masks for the six benchmarked semantic segmentation pipelines in the worst case (held-out test trial 130), using each pipeline's representative seed. The numerical value in each image panel denotes its respective HD95 magnitude (pixels).

During mid-opening, the from-scratch Swin-UNet oversegmented the entire midline opening (75.48 pixels). This oversegmentation was less pronounced but still present for the UNet and UNet++. Both TransUNet pipelines and the pretrained Swin-UNet showed smaller amounts of oversegmentation at the midline, although the from-scratch TransUNet also undersegmented the upper corner of the left vocal fold. Strong residual laser glare was visible near the top of the vocal folds in this phase. In the fully open phase, all pipelines moderately undersegmented along the inner edges of the vocal folds (16.04–23.14 pixels). Tissue-to-flow contrast was relatively low in this region and phase. At closing, all pipelines closely matched the ground truth (5.83–12.04 pixels), with the pretrained Swin-UNet mildly oversegmenting the inner edge of the vocal folds.

In addition to the best- and worst-case visualizations of trials 42 and 130, trial 105 was selected because it showed the largest reduction in trial-level mean HD95 from pretraining for both the TransUNet and Swin-UNet, as depicted in Fig. 8. The selection was based on the smaller of the two reductions. Because the effect of pretraining is measured as a difference between pipelines, this case was identified using the trial mean rather than the per-image HD95. In trial 105, pretraining reduced the trial-level mean HD95 by 25.00 pixels for the TransUNet and by 35.73 pixels for the Swin-UNet. The number of phase-resolved images with an HD95 $>$ 50 pixels also fell, from 25 to 3 for the TransUNet and from 34 to 4 for the Swin-UNet.

In the closed phase, both the from-scratch and pretrained TransUNet undersegmented the midline of the vocal folds. Unlike in trials 42 and 130, the ground-truth mask showed no slit at the midline in this phase. The from-scratch TransUNet undersegmented a wide region at the midline (112.00 pixels), whereas the pretrained TransUNet predicted a narrow slit (86.00 pixels). In this phase, the other four pipelines aligned closely with the ground truth. During early opening, a slit was visible in the ground-truth mask, extending all the way down the vocal folds. The from-scratch TransUNet and from-scratch Swin-UNet oversegmented nearly the entire slit (97.00 and 105.05 pixels), whereas their pretrained counterparts left much of it open (8.06 and 12.04 pixels). The UNet also oversegmented most of the slit, and the UNet++ its lower part.

At mid-opening, small amounts of oversegmentation were present near the bottom of the opening in the UNet and UNet++ and at the top of the vocal folds in all pipelines except the pretrained TransUNet, which undersegmented there. During the fully open and closing phases, all pipelines closely matched the ground truth, with small amounts of oversegmentation along the left edge of the opening in the fully open phase and undersegmentation at the top of the vocal folds at closing. The UNet and UNet++ also oversegmented the lower part of the opening during closing. In these three phases, pretraining did not consistently lower HD95, and the pretrained variant had the higher value in two of the six comparisons, the Swin-UNet at mid-opening (16.16 vs 14.21 pixels) and the TransUNet at closing (7.81 vs 6.14 pixels).

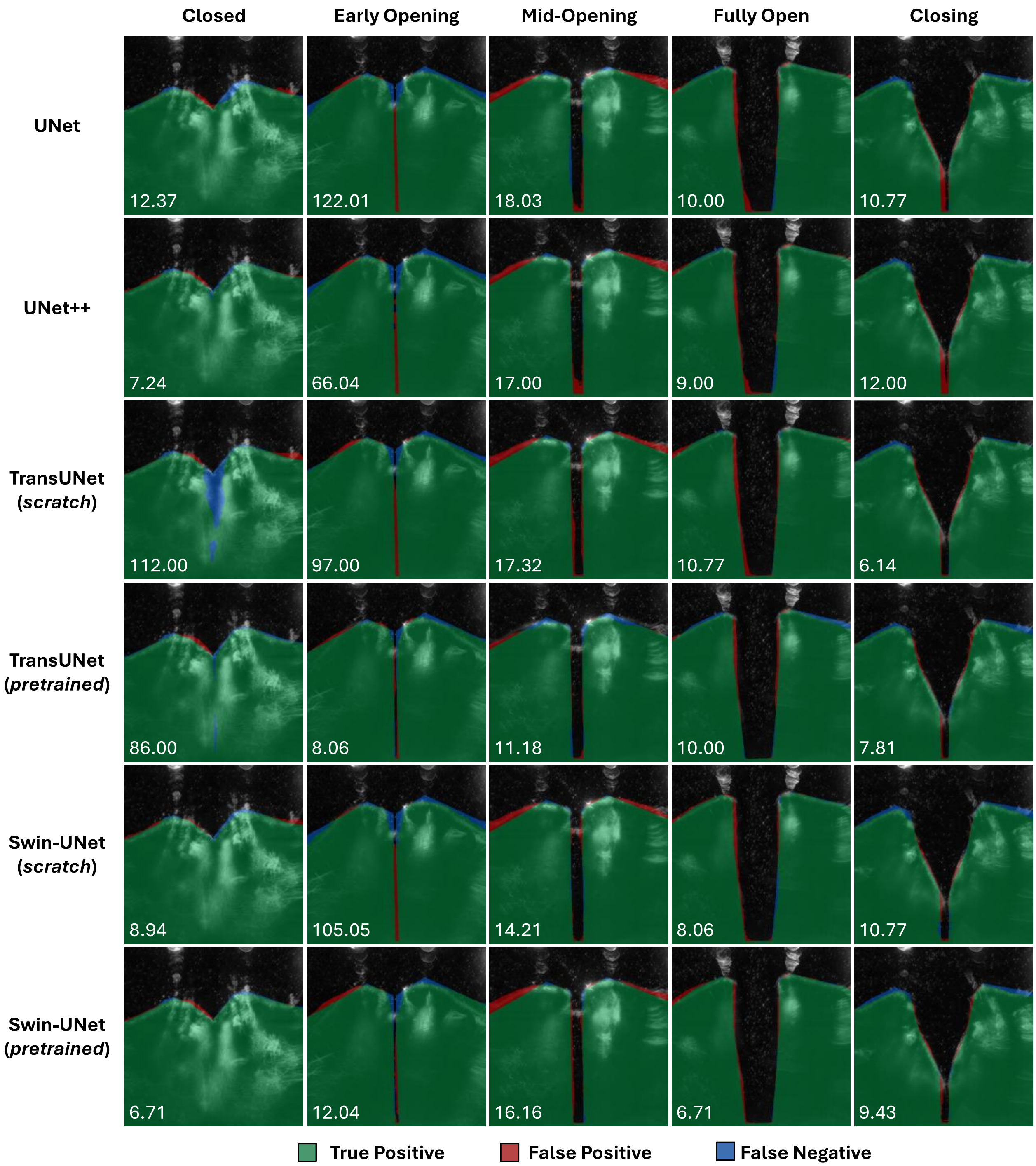

**Fig. 8.** Predicted masks overlaid on the ground-truth masks for the six benchmarked semantic segmentation pipelines in the trial with the largest reduction in HD95 from pretraining (held-out test trial 105), using each pipeline's representative seed. The numerical value in each image panel denotes its respective HD95 magnitude (pixels).

## 4. Discussion

### 4.1 Effect of pretraining

The hybrid CNN-transformer-based TransUNet and the pure transformer-based Swin-UNet were trained from scratch and with ImageNet-pretrained initialization to assess the effect of large-scale pretraining on vocal fold segmentation in 2D intraglottal PIV high-speed images. For both architectures, statistical testing showed that pretraining was associated with a significantly higher DSC and BF1 and a significantly lower HD95. For the three metrics, all seed-specific medians agreed in direction. The DSC differences were small (0.0010–0.0024), so pretraining mainly improved boundary fidelity rather than regional overlap. This improvement was considerably larger for the Swin-UNet, whose median paired differences in BF1 and HD95 were about four times those of the TransUNet. The TransUNet HD95 contrast had the largest $p$ value of the seven significant results and should be interpreted with caution.

On the held-out test set, pretraining improved mean BF1 by 0.0254 and reduced mean HD95 by 7.22 pixels for the Swin-UNet, compared with 0.0076 and 1.40 pixels for the TransUNet. Given the 2-pixel boundary tolerance, these BF1 gains reflect closer agreement between the predicted and ground-truth vocal fold boundaries. For HD95, however, the benefit came mainly from the most difficult images. For the Swin-UNet, pretraining lowered the 90$^{th}$ percentile of per-image HD95 values from 80.4 pixels to 52.0 pixels and the proportion of images with HD95 $>$ 50 pixels from 15.8% to 10.3%, whereas the median decreased only from 19.4 to 15.2 pixels. For the TransUNet, the corresponding changes were smaller, from 57.6 to 52.8 pixels, from 11.6% to 10.5%, and from 17.0 to 15.8 pixels.

This asymmetry is consistent with the inductive biases of the two architectures. Unlike convolution, self-attention does not embed locality and translation equivariance into each layer. As a result, a transformer trained from scratch must learn much of this spatial structure from the training data, which underlies its data dependency [26]. The Swin-UNet is a pure transformer-based hierarchical network, with transformer blocks in both its encoder and decoder [29]. Although its shifted windows restrict self-attention to local regions [39], the from-scratch Swin-UNet must learn low-level features such as edges and boundary continuity without a CNN encoder to provide them. Large-scale pretraining has been demonstrated to help the lower attention layers of ViTs incorporate local information [48], and to help them perform on par with CNN-based models for medical image classification [32]. A similar pattern emerged in this study, as pretraining moved the Swin-UNet from the lowest mean DSC and BF1 of the six semantic segmentation pipelines to the highest on the held-out test set.

The TransUNet implementation in this work passes each image through a ResNetV2 CNN encoder before the transformer [38] and retains its intermediate feature maps as skip connections to the CUP decoder [28], preserving the locality and translation equivariance of convolution without pretraining. Consistent with this, the from-scratch TransUNet had a higher mean DSC (0.9740 vs 0.9734) and BF1 (0.6594 vs 0.6425) and a lower mean HD95 (26.42 vs 30.83 pixels) than the from-scratch Swin-UNet when evaluated on the held-out test set. Additionally, the from-scratch TransUNet had about four times as many trainable parameters as the from-scratch Swin-UNet. However, this larger capacity is unlikely to fully explain the difference because larger transformers need more data to realize their full performance, as larger ViT models underperformed smaller ones even when pretrained on ImageNet [26].

Notably, both pretrained checkpoints were trained on natural color images, yet they improved the segmentation of grayscale intraglottal PIV images mainly at the vocal fold boundary. Networks trained on natural images learn first-layer features resembling Gabor filters that are general across datasets and tasks [49]. Furthermore, the patch-embedding filters of ViTs resemble basis functions for the fine structure within each patch [26]. Such edge-sensitive features may have been most useful at the tissue-to-flow interface, where the optical conditions of intraglottal PIV obscure the vocal fold boundaries. The value of ImageNet pretraining for this task may therefore lie in these general low-level features rather than in the content of the pretraining images.

The pretraining contrasts in this study differed in more than their initial weights. The Swin-UNet checkpoint initialized both its encoder and decoder transformer stages, whereas the CUP decoder of the pretrained TransUNet was trained from scratch. This could have enlarged the effect of pretraining for the Swin-UNet. Each pretrained pipeline also had its own selected peak LR, training duration, and three-channel input. In a recent histopathological benchmark, in which the decoders of all pretrained models were initialized randomly, the TransUNet instead gained more from pretraining than the Swin-UNet in DSC and BF1 [33]. Differences in pretrained checkpoints, decoder initialization, LR selection, and segmentation task could all account for this difference. This suggests that the benefit of pretraining depends on the task and implementation, not only on the architecture family. For vocal fold segmentation in 2D intraglottal PIV high-speed images, the results support using the TransUNet and Swin-UNet with ImageNet-pretrained initialization, particularly the Swin-UNet.

### 4.2 Effect of architecture

Without pretraining, the CNN-based UNet and UNet++ were each compared with the TransUNet and Swin-UNet in four exploratory *post hoc* contrasts to assess the effect of architecture on vocal fold segmentation in 2D intraglottal PIV high-speed images. Statistical testing showed that the from-scratch TransUNet had a significantly higher BF1 and a significantly lower HD95 than the UNet, with median paired differences of 0.0100 and 6.30 pixels. For these two metrics, all seed-specific medians agreed in direction. Against the UNet++, these differences were also significant, but their CIs included zero and the seed-specific medians did not all agree for BF1. Therefore, this advantage should be interpreted with caution. The DSC differences were small (0.0005 and 0.0009), and only the higher DSC of the UNet++ was significant, so the benefit of the from-scratch TransUNet was mainly in boundary fidelity rather than regional overlap. Conversely, the UNet and UNet++ had a significantly higher DSC and BF1 than the from-scratch Swin-UNet, and all seed-specific medians agreed in direction, whereas the HD95 differences were not significant.

As with pretraining, the HD95 advantage of the from-scratch TransUNet mainly came from the most difficult images. Compared with the UNet and UNet++, it had a lower 90$^{th}$ percentile of per-image HD95 values (57.6 vs 80.0 and 79.8 pixels) and about half the proportion of images with HD95 $>$ 50 pixels (11.6% vs 22.0% and 21.3%), close to those of the pretrained TransUNet and Swin-UNet. The from-scratch Swin-UNet also had fewer images with HD95 $>$ 50 pixels (15.8%), but its 90$^{th}$ percentile of 80.4 pixels was similar to those of the UNet and UNet++.

These contrasts are consistent with the trade-off between the locality of convolution and the long-range context of self-attention. The effective receptive field of a CNN grows much more slowly with network depth than its theoretical receptive field [25], which may limit the context available to the UNet and UNet++ in the most difficult images. In the TransUNet, self-attention operates globally on the tokenized CNN feature map [28], and even early transformer layers aggregate more global information than convolutional layers at the same depth [48]. The Swin-UNet instead restricts self-attention to shifted local windows [29,39], which may explain why the from-scratch Swin-UNet had a 90$^{th}$ percentile similar to those of the UNet and UNet++. Similarly, in the histopathological benchmark, the from-scratch TransUNet had a higher mean DSC and BF1 than the from-scratch Swin-UNet on both of its datasets [33]. However, the TransUNet has about three times as many trainable parameters as the UNet and UNet++, so its advantage cannot be attributed to self-attention alone. The from-scratch Swin-UNet, in contrast, has fewer trainable parameters than either CNN-based pipeline, yet the pretrained Swin-UNet, with nearly the same number of parameters, had the highest mean DSC and BF1 of the six pipelines on the held-out test set. Its lower DSC and BF1 therefore point to the lack of low-level features from a CNN encoder or pretraining rather than to limited capacity.

Once pretrained, the difference between the TransUNet and Swin-UNet largely disappeared. In the prespecified top-two contrast, the pretrained Swin-UNet had a significantly higher DSC than the pretrained TransUNet, but the median paired difference was only 0.0010, and the BF1 and HD95 differences were not significant. Furthermore, their per-image HD95 distributions nearly coincided. Both pretrained architectures had a higher mean DSC and BF1 and a lower mean HD95 than the UNet and UNet++. Therefore, the choice between a hybrid CNN-transformer- and a pure transformer-based architecture mattered mainly without pretraining. For vocal fold segmentation in 2D intraglottal PIV high-speed images, the results support the TransUNet when training from scratch, whereas once pretrained, the Swin-UNet performed similarly with about a quarter of the trainable parameters.

### 4.3 Implications for automated vocal fold segmentation

Fundamental insights into healthy phonation [5,9] and treatment-relevant insights for unilateral vocal fold paralysis [12,13] have been derived from 2D intraglottal PIV. The high-speed images from these *ex vivo* phonation experiments are used to quantify vocal fold medial surface dynamics and the intraglottal flow field along with their interactions. These analyses depend on accurate vocal fold delineation, because flow features such as intraglottal flow separation form along the medial surface and aid in characterizing healthy and pathological phonation [5,34]. Boundary fidelity, more so than regional overlap, is therefore the more relevant measure of segmentation performance for this application.

On the held-out test set, mean DSC differed by only 0.0030 across the six semantic segmentation pipelines, and no significant DSC difference had a median paired difference above 0.0024. All pipelines therefore achieved high regional overlap with the ground-truth mask. However, because most foreground pixels in the mask lie away from the vocal fold boundary, boundary errors do not substantially affect the DSC. Distance-based metrics are recommended when boundary fidelity is important [43]. Mean BF1 differentiated the pipelines more, but even the highest value (0.6679) indicates that the predicted and ground-truth boundaries often differed by more than 2 pixels. HD95 showed how large these discrepancies became. In most images they were moderate, with median per-image HD95 values of 15.2–23.5 pixels. However, even the pretrained TransUNet and Swin-UNet had HD95 $>$ 50 pixels in about 10% of images, compared with more than 20% for the UNet and UNet++.

Where these large errors occurred is just as important as how often they occurred. In the qualitative segmentation overlays, every panel with HD95 $>$ 50 pixels was in the closed, early opening, or mid-opening phase, mostly at the midline where the vocal folds coapt or begin to separate. At closing, HD95 did not exceed 12.04 pixels for any pipeline. Agreement was therefore highest in the phase most relevant to flow separation, in which the glottis is divergent [5–7] and annotation was focused. The largest discrepancies may partly reflect the ground-truth mask rather than segmentation failure. For 2D intraglottal PIV, pretrained Swin-UNet and TransUNet could automate the segmentation pipeline with close alignment to the ground truth, with SME review focused on the midline in the closed and opening phases.

### 4.4 Limitations and future work

The ground-truth masks used in this work were created for PIV analysis, with intermittent annotation and linear interpolation outside the divergent phases and re-annotation of masks with considerable errors. The reported metrics measure agreement with the reference masks used in practice rather than absolute anatomical accuracy. All semantic segmentation pipelines were trained and evaluated against the same masks with paired tests, so the comparisons of architecture and pretraining should be less affected than the absolute values. The trials were also partitioned within each experiment, so the held-out test set contained unseen trials rather than unseen larynges. Although each trial used different experimental conditions, such as subglottal pressure, trials from the same larynx may share specimen-specific features, and the statistical tests did not model this correlation. All images came from *ex vivo* canine larynges, so generalization to new specimens such as cadaveric human larynges remains to be established.

Several design choices favored a controlled comparison over the highest attainable performance. The UNet and UNet++ were trained only from scratch, because pretraining them would have required replacing their encoders with a pretrained backbone. In a recent benchmark, pretraining benefited CNN-based architectures, although less than hybrid CNN-transformer- and pure transformer-based architectures [33]. Whether a pretrained UNet and UNet++ would match the pretrained TransUNet and Swin-UNet remains untested. Self-configuring frameworks such as nnUNet [41] were excluded because they adapt the architecture and training to each dataset, which would confound the effects being compared in this study. No data augmentation was used since the vocal folds are in the same orientation in this experimental framework. The selected peak LR of the pretrained TransUNet was the lowest value in the candidate LR grid. Its optimum may lie below the tested range. The from-scratch TransUNet and Swin-UNet were not compared directly, and the phases in which large errors occurred were identified from three trials at each pipeline's representative seed. Finally, each image was segmented independently, without using the continuity of the 72 phase-resolved images of each vibratory cycle.

Future work will focus on improving both the masks and the pipelines using an iterative human-in-the-loop approach [50]. The trained pipelines could segment the vocal folds in new *ex vivo* phonation experiments, after which an SME corrects the images with the largest errors. The corrected trials are then added to the training set, progressively improving the masks and the segmentation performance. A test set of images from experiments with unseen larynges from a new specimen would show whether the pipelines generalize. A subset of images annotated by several SMEs would show how closely experts agree with each other, and thus the level of agreement the pipelines can be expected to reach. Repeating the PIV analysis with the predicted and manually generated masks would show how the remaining boundary errors affect the intraglottal flow field, and deep learning models that exploit temporal continuity may reduce the segmentation errors at the midline near closure.

## 5. Conclusions

This study assessed the effects of deep learning architecture and ImageNet pretraining on automated vocal fold segmentation in 2D intraglottal PIV high-speed images. Six semantic segmentation pipelines derived from four established architectures spanning three categories were benchmarked under a standardized protocol. Regional overlap varied little across pipelines, and both architecture and pretraining mainly affected boundary fidelity. ImageNet pretraining improved both the TransUNet and Swin-UNet, with the Swin-UNet benefiting more. Once pretrained, the two architectures performed similarly, although the Swin-UNet has only about a quarter of the trainable parameters. Without pretraining, the hybrid CNN-transformer-based TransUNet is preferable, whereas the pure transformer-based Swin-UNet had a lower DSC and BF1 than the CNN-based UNet and UNet++. To our knowledge, this is the first study to apply deep learning to vocal fold segmentation in 2D intraglottal PIV high-speed images. More broadly, this work offers a reference framework for segmenting experimental flow-visualization images with noisy optical conditions or highly deformable objects.

## Data availability

The raw image data and code supporting the conclusions made in this study will be made available by the authors upon request.

## Declaration of competing interest

The authors declare that they have no known competing personal or financial interests that could have appeared to influence the work reported in this manuscript.

## References

[1] Z. Zhang, Mechanics of human voice production and control, The Journal of the Acoustical Society of America 140 (2016) 2614–2635. https://doi.org/10.1121/1.4964509.

[2] G. Fant, Acoustic Theory of Speech Production: With Calculations based on X-Ray Studies of Russian Articulations, DE GRUYTER, 1971. https://doi.org/10.1515/9783110873429.

[3] I.R. Titze, Vocal efficiency, Journal of Voice 6 (1992) 135–138. https://doi.org/10.1016/S0892-1997(05)80127-4.

[4] S.L. Thomson, L. Mongeau, S.H. Frankel, Aerodynamic transfer of energy to the vocal folds, The Journal of the Acoustical Society of America 118 (2005) 1689–1700. https://doi.org/10.1121/1.2000787.

[5] L. Oren, E. Gutmark, S. Khosla, Intraglottal velocity and pressure measurements in a hemilarynx model, The Journal of the Acoustical Society of America 137 (2015) 935–943. https://doi.org/10.1121/1.4906833.

[6] C. Farbos De Luzan, L. Oren, E. Gutmark, S.M. Khosla, Quantification of the Intraglottal Pressure Induced by Flow Separation Vortices Using Large Eddy Simulation, Journal of Voice 35 (2021) 822–831. https://doi.org/10.1016/j.jvoice.2020.02.013.

[7] E. Sundström, L. Oren, C. Farbos De Luzan, E. Gutmark, S. Khosla, Fluid-Structure Interaction Analysis of Aerodynamic and Elasticity Forces During Vocal Fold Vibration, Journal of Voice 39 (2025) 293–303. https://doi.org/10.1016/j.jvoice.2022.08.030.

[8] E. Kirsh, S.R.C. Zacharias, A. De Alarcon, D. Deliyski, M. Tabangin, S. Khosla, Vertical Phase Difference and Glottal Efficiency in Musical Theater and Opera Singers, Journal of Voice 31 (2017) 130.e19-130.e25. https://doi.org/10.1016/j.jvoice.2015.11.025.

[9] J. Michaud-Dorko, C. Farbos De Luzan, G.R. Dion, E. Gutmark, L. Oren, Comparison of Aerodynamic and Elastic Properties in Tissue and Synthetic Models of Vocal Fold Vibrations, Bioengineering 11 (2024) 834. https://doi.org/10.3390/bioengineering11080834.

[10] S. Scharnowski, C.J. Kähler, Particle image velocimetry - Classical operating rules from today's perspective, Optics and Lasers in Engineering 135 (2020) 106185. https://doi.org/10.1016/j.optlaseng.2020.106185.

[11] L. Oren, S. Khosla, E. Gutmark, Medial Surface Dynamics as a Function of Subglottal Pressure in a Canine Larynx Model, Journal of Voice 35 (2021) 69–76. https://doi.org/10.1016/j.jvoice.2019.07.015.

[12] A. Maddox, L. Oren, C. Farbos De Luzan, R. Howell, E. Gutmark, S. Khosla, An Ex-vivo Model Examining Acoustics and Aerodynamic Effects Following Medialization With and Without Arytenoid Adduction, The Laryngoscope 133 (2023) 621–627. https://doi.org/10.1002/lary.30235.

[13] L. Oren, A. Maddox, C. Farbos De Luzan, C. Xie, R. Howell, G. Dion, E. Gutmark, S. Khosla, Acoustics and aerodynamic effects following glottal and infraglottal medialization in an excised larynx model, Eur Arch Otorhinolaryngol 281 (2024) 2523–2529. https://doi.org/10.1007/s00405-024-08519-x.

[14] A. Masullo, R. Theunissen, Automated mask generation for PIV image analysis based on pixel intensity statistics, Exp Fluids 58 (2017) 70. https://doi.org/10.1007/s00348-017-2357-3.
[15] P. Nikoueeyan, J.W. Naughton, A photogrammetric approach for masking particle image velocimetry images around moving bodies*, Meas. Sci. Technol. 29 (2018) 105203. https://doi.org/10.1088/1361-6501/aad9c8.
[16] V. Lombardi, M.L. Rocca, P. Prestininzi, A new dynamic masking technique for time resolved PIV analysis: A new freeware raycasting tool combined with fluorescent coating to allow for object visibility, J Vis 24 (2021) 979–990. https://doi.org/10.1007/s12650-021-00756-0.
[17] B. Vennemann, T. Rösgen, A dynamic masking technique for particle image velocimetry using convolutional autoencoders, Exp Fluids 61 (2020) 168. https://doi.org/10.1007/s00348-020-02984-w.
[18] G. Andrade-Miranda, K. Chatzipapas, J.D. Arias-Londoño, J.I. Godino-Llorente, GIRAFE: Glottal imaging dataset for advanced segmentation, analysis, and facilitative playbacks evaluation, Data in Brief 59 (2025) 111376. https://doi.org/10.1016/j.dib.2025.111376.
[19] S.M.N. Nobel, S.M.M.R. Swapno, Md.R. Islam, M. Safran, S. Alfarhood, M.F. Mridha, A machine learning approach for vocal fold segmentation and disorder classification based on ensemble method, Sci Rep 14 (2024) 14435. https://doi.org/10.1038/s41598-024-64987-5.
[20] Y. Zhang, Y. Song, J. Liu, M. Li, An automatic laryngoscopic image segmentation system based on SAM prompt engineering: from glottis annotation to vocal fold segmentation, Front. Mol. Biosci. 12 (2025) 1616271. https://doi.org/10.3389/fmolb.2025.1616271.
[21] H. Ding, Q. Cen, X. Si, Z. Pan, X. Chen, Automatic glottis segmentation for laryngeal endoscopic images based on U-Net, Biomedical Signal Processing and Control 71 (2022) 103116. https://doi.org/10.1016/j.bspc.2021.103116.
[22] M.H. Hesamian, W. Jia, X. He, P. Kennedy, Deep Learning Techniques for Medical Image Segmentation: Achievements and Challenges, J Digit Imaging 32 (2019) 582–596. https://doi.org/10.1007/s10278-019-00227-x.
[23] O. Ronneberger, P. Fischer, T. Brox, U-Net: Convolutional Networks for Biomedical Image Segmentation, in: N. Navab, J. Hornegger, W.M. Wells, A.F. Frangi (Eds.), Medical Image Computing and Computer-Assisted Intervention – MICCAI 2015, Springer International Publishing, Cham, 2015: pp. 234–241. https://doi.org/10.1007/978-3-319-24574-4_28.
[24] Z. Zhou, M.M. Rahman Siddiquee, N. Tajbakhsh, J. Liang, UNet++: A Nested U-Net Architecture for Medical Image Segmentation, in: D. Stoyanov, Z. Taylor, G. Carneiro, T. Syeda-Mahmood, A. Martel, L. Maier-Hein, J.M.R.S. Tavares, A. Bradley, J.P. Papa, V. Belagiannis, J.C. Nascimento, Z. Lu, S. Conjeti, M. Moradi, H. Greenspan, A. Madabhushi (Eds.), Deep Learning in Medical Image Analysis and Multimodal Learning for Clinical Decision Support, Springer International Publishing, Cham, 2018: pp. 3–11. https://doi.org/10.1007/978-3-030-00889-5_1.
[25] W. Luo, Y. Li, R. Urtasun, R. Zemel, Understanding the Effective Receptive Field in Deep Convolutional Neural Networks, (2017). https://doi.org/10.48550/ARXIV.1701.04128.

[26] A. Dosovitskiy, L. Beyer, A. Kolesnikov, D. Weissenborn, X. Zhai, T. Unterthiner, M. Dehghani, M. Minderer, G. Heigold, S. Gelly, J. Uszkoreit, N. Houlsby, An Image is Worth 16x16 Words: Transformers for Image Recognition at Scale, (2020). https://doi.org/10.48550/ARXIV.2010.11929.
[27] H. Xiao, L. Li, Q. Liu, X. Zhu, Q. Zhang, Transformers in medical image segmentation: A review, Biomedical Signal Processing and Control 84 (2023) 104791. https://doi.org/10.1016/j.bspc.2023.104791.
[28] J. Chen, Y. Lu, Q. Yu, X. Luo, E. Adeli, Y. Wang, L. Lu, A.L. Yuille, Y. Zhou, TransUNet: Transformers Make Strong Encoders for Medical Image Segmentation, (2021). https://doi.org/10.48550/ARXIV.2102.04306.
[29] H. Cao, Y. Wang, J. Chen, D. Jiang, X. Zhang, Q. Tian, M. Wang, Swin-Unet: Unet-Like Pure Transformer for Medical Image Segmentation, in: L. Karlinsky, T. Michaeli, K. Nishino (Eds.), Computer Vision – ECCV 2022 Workshops, Springer Nature Switzerland, Cham, 2023: pp. 205–218. https://doi.org/10.1007/978-3-031-25066-8_9.
[30] H. Kang, W. Zhang, S. Li, X. Wang, Y. Sun, X. Sun, F.-X. Li, C. Hou, S. Lam, Y. Zheng, A comprehensive benchmarking of a U-Net based model for midbrain auto-segmentation on transcranial sonography, Computer Methods and Programs in Biomedicine 258 (2025) 108494. https://doi.org/10.1016/j.cmpb.2024.108494.
[31] G. Vazquez, D. Ragusa, E. Torti, E. Marenzi, E. Juarez, A.M. Groba, F. Leporati, Benchmarking deep learning architectures for hyperspectral in-vivo brain tumor segmentation, Computer Methods and Programs in Biomedicine 286 (2026) 109571. https://doi.org/10.1016/j.cmpb.2026.109571.
[32] C. Matsoukas, J.F. Haslum, M. Söderberg, K. Smith, Is it Time to Replace CNNs with Transformers for Medical Images?, (2021). https://doi.org/10.48550/ARXIV.2108.09038.
[33] A. Gulsen, Benchmarking Deep Segmentation Architectures for Histopathological Nuclei Segmentation: A Controlled Study of Transfer Learning, Robustness, and Efficiency Across CNN, Transformer, and Hybrid Models, J Digit Imaging. Inform. Med. (2026). https://doi.org/10.1007/s10278-026-02186-6.
[34] J. Michaud-Dorko, C.F.D. Luzan, G.R. Dion, E. Gutmark, L. Oren, Phase-locked characterization of intraglottal flow separation vortices and medial surface motion during phonation, Physics of Fluids 38 (2026) 051911. https://doi.org/10.1063/5.0291498.
[35] L. Oren, S. Khosla, E. Gutmark, Intraglottal geometry and velocity measurements in canine larynges, The Journal of the Acoustical Society of America 135 (2014) 380–388. https://doi.org/10.1121/1.4837222.
[36] S. Kapoor, A. Narayanan, Leakage and the reproducibility crisis in machine-learning-based science, Patterns 4 (2023) 100804. https://doi.org/10.1016/j.patter.2023.100804.
[37] K. He, X. Zhang, S. Ren, J. Sun, Identity Mappings in Deep Residual Networks, in: B. Leibe, J. Matas, N. Sebe, M. Welling (Eds.), Computer Vision – ECCV 2016, Springer International Publishing, Cham, 2016: pp. 630–645. https://doi.org/10.1007/978-3-319-46493-0_38.
[38] A. Kolesnikov, L. Beyer, X. Zhai, J. Puigcerver, J. Yung, S. Gelly, N. Houlsby, Big Transfer (BiT): General Visual Representation Learning, in: A. Vedaldi, H. Bischof, T. Brox, J.-M.

Frahm (Eds.), Computer Vision – ECCV 2020, Springer International Publishing, Cham, 2020: pp. 491–507. https://doi.org/10.1007/978-3-030-58558-7_29.
[39] Z. Liu, Y. Lin, Y. Cao, H. Hu, Y. Wei, Z. Zhang, S. Lin, B. Guo, Swin Transformer: Hierarchical Vision Transformer using Shifted Windows, in: 2021 IEEE/CVF International Conference on Computer Vision (ICCV), IEEE, Montreal, QC, Canada, 2021: pp. 9992–10002. https://doi.org/10.1109/ICCV48922.2021.00986.
[40] I. Goodfellow, Y. Bengio, A. Courville, Deep Learning, The MIT Press, 2016.
[41] F. Isensee, P.F. Jaeger, S.A.A. Kohl, J. Petersen, K.H. Maier-Hein, nnU-Net: a self-configuring method for deep learning-based biomedical image segmentation, Nat Methods 18 (2021) 203–211. https://doi.org/10.1038/s41592-020-01008-z.
[42] G. Csurka, D. Larlus, F. Perronnin, What is a good evaluation measure for semantic segmentation?, in: Procedings of the British Machine Vision Conference 2013, British Machine Vision Association, Bristol, 2013: p. 32.1-32.11. https://doi.org/10.5244/C.27.32.
[43] A.A. Taha, A. Hanbury, Metrics for evaluating 3D medical image segmentation: analysis, selection, and tool, BMC Med Imaging 15 (2015) 29. https://doi.org/10.1186/s12880-015-0068-x.
[44] L. Maier-Hein, A. Reinke, P. Godau, M.D. Tizabi, F. Buettner, E. Christodoulou, B. Glocker, F. Isensee, J. Kleesiek, M. Kozubek, M. Reyes, M.A. Riegler, M. Wiesenfarth, A.E. Kavur, C.H. Sudre, M. Baumgartner, M. Eisenmann, D. Heckmann-Nötzel, T. Rädsch, L. Acion, M. Antonelli, T. Arbel, S. Bakas, A. Benis, M.B. Blaschko, M.J. Cardoso, V. Cheplygina, B.A. Cimini, G.S. Collins, K. Farahani, L. Ferrer, A. Galdran, B. Van Ginneken, R. Haase, D.A. Hashimoto, M.M. Hoffman, M. Huisman, P. Jannin, C.E. Kahn, D. Kainmueller, B. Kainz, A. Karargyris, A. Karthikesalingam, F. Kofler, A. Kopp-Schneider, A. Kreshuk, T. Kurc, B.A. Landman, G. Litjens, A. Madani, K. Maier-Hein, A.L. Martel, P. Mattson, E. Meijering, B. Menze, K.G.M. Moons, H. Müller, B. Nichyporuk, F. Nickel, J. Petersen, N. Rajpoot, N. Rieke, J. Saez-Rodriguez, C.I. Sánchez, S. Shetty, M. Van Smeden, R.M. Summers, A.A. Taha, A. Tiulpin, S.A. Tsaftaris, B. Van Calster, G. Varoquaux, P.F. Jäger, Metrics reloaded: recommendations for image analysis validation, Nat Methods 21 (2024) 195–212. https://doi.org/10.1038/s41592-023-02151-z.
[45] M. Wiesenfarth, A. Reinke, B.A. Landman, M. Eisenmann, L.A. Saiz, M.J. Cardoso, L. Maier-Hein, A. Kopp-Schneider, Methods and open-source toolkit for analyzing and visualizing challenge results, Sci Rep 11 (2021) 2369. https://doi.org/10.1038/s41598-021-82017-6.
[46] S. Holm, A Simple Sequentially Rejective Multiple Test Procedure, Scandinavian Journal of Statistics 6 (1979) 65–70.
[47] Z. Huang, J. Ye, H. Wang, Z. Deng, Z. Yang, Y. Su, J. Liu, T. Li, Y. Gu, S. Zhang, Y. Qiao, L. Gu, J. He, Revisiting model scaling with a U-net benchmark for 3D medical image segmentation, Sci Rep 15 (2025) 29795. https://doi.org/10.1038/s41598-025-15617-1.
[48] M. Raghu, T. Unterthiner, S. Kornblith, C. Zhang, A. Dosovitskiy, Do Vision Transformers See Like Convolutional Neural Networks?, (2021). https://doi.org/10.48550/ARXIV.2108.08810.

[49] J. Yosinski, J. Clune, Y. Bengio, H. Lipson, How transferable are features in deep neural networks?, (2014). https://doi.org/10.48550/ARXIV.1411.1792.
[50] V.V. Kalaiarasan, M. Miller, X. Han, B. Foreman, X. Jia, A Novel Methodology for Intracranial Pressure Subpeak Identification Enabling Morphological Feature Analysis, IEEE Trans. Biomed. Eng. 72 (2025) 1288–1297. https://doi.org/10.1109/TBME.2024.3495542.